%% file: iclr2025_conference.tex
\documentclass{article} % For LaTeX2e
\usepackage{iclr2025_conference,times}

\input{math_commands.tex}

\usepackage{hyperref}
\usepackage{url}

\usepackage{multicol}
\usepackage{amssymb}
\usepackage{bm}
\usepackage{amsmath}
\usepackage{graphicx}
\usepackage{xcolor}
\usepackage{import}
\usepackage{mathtools}
\usepackage{macros}
\usepackage{svg}
\usepackage{soul}
\usepackage{makecell}
\usepackage{multirow}
\usepackage{array}
\usepackage{colortbl}
\usepackage{siunitx}
\usepackage{adjustbox}
\usepackage{subcaption}
\usepackage{adjustbox}
\usepackage{booktabs}
\usepackage[nolist]{acronym}

\title{A Riemannian Framework for Learning \\ Reduced-order Lagrangian Dynamics}

\author{Katharina Friedl\textsuperscript{\ensuremath{1}} \quad No\'emie Jaquier\textsuperscript{\ensuremath{1,2}} \quad Jens Lundell\textsuperscript{\ensuremath{1}} \quad Tamim Asfour\textsuperscript{\ensuremath{2}} \quad Danica Kragic\textsuperscript{\ensuremath{1}} \\
\textsuperscript{\ensuremath{1}} Division of Robotics, Perception, and Learning \quad \textsuperscript{\ensuremath{2}} Institute for Anthropomatics and Robotics\\
\hspace{1.0cm} KTH Royal Institute of Technology \hspace{1.4cm} Karlsruhe Institute of Technology (KIT)\\
\quad\quad\quad\quad \texttt{\small \{kfriedl,jaquier,jelundel,dani\}@kth.se, asfour@kit.edu} \\
}

\iclrfinalcopy % Uncomment for camera-ready version, but NOT for submission.
\begin{document}

\input{acronyms}

\maketitle

\begin{abstract}
\input{texfiles/00abstract.tex}
\end{abstract}

\input{texfiles/01introduction.tex}

\input{texfiles/02preliminaries.tex}
% network architecture
\input{texfiles/03method.tex}

% results
\input{texfiles/04results.tex}

% conclusion
\input{texfiles/05conclusion.tex}

\clearpage
\subsubsection*{Acknowledgments}
This work was supported by ERC AdV grant BIRD, Knut and Alice Wallenberg Foundation and Swedish Research Council.

\bibliography{references}
\bibliographystyle{iclr2025_conference}

\clearpage
\appendix
\input{texfiles/technical_appendix.tex}
\input{texfiles/experimental_appendix.tex}

\end{document}

%% file: math_commands.tex
\usepackage{amsmath,amsfonts,bm}

\def\1{\bm{1}}

\DeclareMathAlphabet{\mathsfit}{\encodingdefault}{\sfdefault}{m}{sl}
\SetMathAlphabet{\mathsfit}{bold}{\encodingdefault}{\sfdefault}{bx}{n}

%% file: acronyms.tex
\newacro{ae}[AE]{Autoencoder}
\newacro{rom}[ROM]{reduced-order model}
\newacro{fom}[FOM]{full-order model}
\newacro{mor}[MOR]{model order reduction}
\newacro{lnn}[LNN]{Lagrangian neural network}
\newacro{dof}[DoF]{degrees-of-freedom}
\newacro{ivp}[IVP]{initial value problem}
\newacro{dnn}[DNN]{Deep Neural Network}
\newacro{spd}[SPD]{symmetric positive-definite}
\newacro{mlp}[MLP]{multilayer perceptron}
\newacro{fc}[FC]{fully-connected}
\newacro{gyrospd}[GyroSpd$_{\ty{++}}$]{gyrospace hyperplane-based}
\newacro{gyroai}[GyroAI]{gyrocalculus-based}
\newacro{rolnn}[RO-LNN]{reduced-order LNN}

%% file: texfiles/00abstract.tex
\vspace{-0.2cm}
\label{abstract}
By incorporating physical consistency as inductive bias, deep neural networks display increased generalization capabilities and data efficiency in learning nonlinear dynamic models. However, the complexity of these models generally increases with the system dimensionality, requiring larger datasets, more complex deep networks, and significant computational effort.
We propose a novel geometric network architecture to learn physically-consistent reduced-order dynamic parameters that accurately describe the original high-dimensional system behavior.
This is achieved by building on recent advances in model-order reduction and by adopting a Riemannian perspective to jointly learn a non-linear structure-preserving latent space and the associated low-dimensional dynamics.
Our approach enables accurate long-term predictions of the high-dimensional dynamics of rigid and deformable systems with increased data efficiency by inferring interpretable and physically-plausible reduced Lagrangian models.
\vspace{-0.2cm}

%% file: texfiles/01introduction.tex
\vspace{-0.3cm}
\section{Introduction}
\label{sec:introduction}
\vspace{-0.2cm}
Deep learning models recently emerged as powerful tools for learning the continuous-time dynamics of physical systems. In contrast to classical system identification methods which rely on analytical derivations, black-box approaches learn to predict the behavior of nonlinear systems using large amounts of trajectory data.
However, collecting such data is prohibitively expensive, and the learned models often predict trajectories that violate the laws of physics, e.g., by not conserving energy.

Gray-box approaches tackle such limitations by incorporating physics priors as inductive biases into deep learning architectures. These methods have been broadly investigated for different systems, such as Lagrangian and Hamiltonian mechanics, and function approximators, including neural networks~\citep{Lutter19:DeLan,Greydanus2019HNN,CranmerGreydanus2020LNN,Lutter2023DeLaN} and Gaussian processes~\citep{Tanaka2022GPs,Evangelisti2022LGP}.
Several physically-consistent models were extended to capture non-conservative forces and contact mechanics~\citep{Hochlehnert2021contact,Zhong2021Contact}, and to ensure longer-term stability of the identified dynamics by including differential solvers~\citep{Chen2018NODE} into the training loop~\citep{Finzi2020, Zhong2020Unsupervised}. Model performances were also improved by representing the dynamics using coordinates adapted to the physical system at hand~\citep{Finzi2020,Celledoni2023HNN,Duong21Hamiltonian}. Altogether, gray-box approaches are physically consistent, versatile, more data-efficient, and display increased generalization capabilities compared to black-box methods~\citep{Lutter2023DeLaN,Greydanus2019HNN}.
Still, they have, so far, only learned the dynamics of low-dimensional systems, i.e., typically $2$-$5$ dimensions. Learning the dynamics of high-dimensional systems, such as fluid flows, continuum mechanics, and robots, arguably remains an open problem. This is due to their increasing complexity, requiring more complex network architectures and more training data.

Predicting trajectories of high-dimensional systems is also notoriously difficult due to the computational cost of solving high-dimensional and highly nonlinear differential equations. In this context, \ac{mor} techniques find a computationally efficient yet descriptive low-dimensional surrogate system --- a \ac{rom} --- of a given high-dimensional dynamical system or \ac{fom} with known dynamics~\citep{Schilders2008MOR}. \ac{mor} is typically achieved by projecting the \ac{fom} onto a lower-dimensional space via linear~\citep{Thieffry2019, Farhat2015StructurepreservingSA, Carlberg2015LagrMOR} or nonlinear~\citep{Sharma2023symplMOR, Barnett2022quadrproj} mappings. 
In particular, \acp{ae} were shown to be well suited to extract descriptive nonlinear reduced representations from data~\citep{LeeCarlberg2020MOR,Buchfink23:SymplecticMOR,Otto2023MOR}.
Recent approaches investigated the preservation of physics-induced geometric structures, i.e. Lagrangian or Hamiltonian structures, of \acp{fom} within projection-based \acp{rom}~\citep{Carlberg2015LagrMOR,LeeCarlberg2020MOR,Otto2023MOR, Hesthaven2021}. This structure awareness enables the derivation of shared theoretical properties such as energy conservation and stability preservation~\citep{Buchfink23:SymplecticMOR}, as well as the design of low-dimensional control strategies on the \ac{rom}~\citep{Lepri2024MOR}.
\citet{Buchfink2024MOR} recently unified various structure-preserving \ac{mor} techniques by adopting a differential-geometric perspective on the problem. In this framework, \acp{rom} are defined on embedded submanifolds of the high-dimensional Riemannian or symplectic manifold associated with a given Lagrangian or Hamiltonian system.

Despite their advantages, MOR methods are intrusive in that they assume known high-dimensional dynamic parameters which are generally difficult to obtain. 
\citet{sharma24} presented a novel non-intrusive approach that identifies low-dimensional dynamic parameters in a structure-preserving linear subspace obtained from high-dimensional state observations. However, this approach is limited to systems that are linearly reducible and displays limited expressivity by constraining the range of the ROM parameters.
\emph{In this paper}, we propose a more general and expressive approach in the form of a physics-inspired geometric deep network that learns the continuous-time dynamics of high-dimensional systems. 
Our \emph{first contribution} is to adopt a differential geometry perspective and jointly learn \emph{(1)} a structure-preserving non-linear reduced representation of the high-dimensional generalized coordinates and \emph{(2)} the associated low-dimensional dynamic parameters. 
Specifically, our approach leverages a constrained \ac{ae}~\citep{Otto2023MOR} to learn a latent embedded submanifold in which physically-consistent low-dimensional dynamics are learned with a \ac{lnn}~\citep{CranmerGreydanus2020LNN,Lutter2023DeLaN}.
Our approach differs from~\citep{Greydanus2019HNN,Zhong2020Unsupervised,Botev24:WhichPriorMatter} in that it does not consider high-dimensional observations (images) of low-dimensional physical systems but systems with high-dimensional state spaces.
As \emph{second} and \emph{third contributions}, we reformulate both the \ac{lnn} and the \ac{ae} to account for the intrinsic geometry of their parameters. To this end, we introduce positive-definite layers in the \ac{lnn} to parameterize the (reduced) mass-inertia matrix and use Riemannian optimization to infer the \ac{lnn} parameters and biorthogonal \ac{ae} weights.

Our approach infers physically-plausible models as both the \ac{rom} and the \ac{fom} conserve the reduced energy along reduced unforced Lagrangian trajectories and their embeddings. Moreover, it infers interpretable reduced-order dynamic parameters. We validate our approach by learning the dynamics of three simulated high-dimensional rigid and deformable systems: a pendulum, a rope, and a thin cloth. 
Our results demonstrate that it efficiently learns reduced-order dynamics leading to accurate long-term predictions of high-dimensional systems.

%% file: texfiles/02preliminaries.tex
\vspace{-0.2cm}
\section{Background}
\label{sec:preliminaries}
\vspace{-0.2cm}
This section introduces the mathematical preliminaries and network architectures for learning reduced Lagrangian dynamics. 
We adopt a differential-geometric perspective 
to simultaneously investigate the use of nonlinear \ac{mor} projections and the structure preservation of Lagrangian systems. 

\vspace{-0.1cm}
\subsection{Lagrangian Dynamics on the Configuration Manifold}
\label{subsec:lagrangiandynamics}
\vspace{-0.2cm}
We consider an $n$-\ac{dof} mechanical system whose configuration space is identified with an $n$-dimensional smooth 
manifold $\mathcalq$ with a simple global chart.
Velocities $\dbmq$ at each configuration $\bmq\in\mathcalq$ lie in the tangent space $\tangentq{\bmq}$, an $n$-dimensional vector space composed of all vectors tangent to $\mathcalq$ at $\bmq$.
The disjoint union of all tangent spaces $\tangentq{\bmq}$ forms the tangent bundle $\tangentbundleq$, a smooth $2n$-dimensional manifold. 
The configuration manifold can be equipped with a Riemannian metric, i.e., a smoothly-varying
inner product acting on $\tangentbundleq$. We consider the kinetic-energy metric, which, given a choice of local coordinates, equals the system's mass-inertia matrix $\massmatrix$.

A Lagrangian system is a tuple $(\mathcalq, \lagrangian)$ of a Riemannian configuration manifold $\mathcalq$ and a smooth time-independent Lagrangian function $\lagrangian: T\mathcalq \to \mathbb{R}$.  
The Lagrangian function is given by the difference between the system's kinetic $\kineticenergy$ and potential $\potentialenergy$ energies as $\lagrangian(\bmq, \dbmq) = \kineticenergy - \potentialenergy = \frac{1}{2}\dbmq^{\intercal} \massmatrix \dbmq - \potentialenergy$.
Following the principle of least action, the equations of motion are given by the Euler-Lagrange equations
$\ddt\big(\dpartial{\lagrangian}{\dbmq}\big) - \dpartial{\lagrangian}{\bmq} = \bmtau$
with generalized forces $\bmtau$. These equations are
\begin{equation}
\label{eq:equationsofmotion}
    \massmatrix\ddbmq + \coriolisterm + \gravityterm = \bmtau,
    \quad\text{with}\quad
    \bm{c}(\bmq, \dbmq) = \big(\ddt\bmM(\bmq)\big)\dbmq - \frac{1}{2}\big(\dfracpartial{\bmq}\left(\dbmq^{\intercal}\bmM(\bmq)\dot{\bmq}\right)\big)^\intercal,
\end{equation}
and $\gravityterm = \dpartial{\potentialenergy}{\bmq}$, where $\coriolisterm$ represents the influence of Coriolis forces. 
Given a time interval $\mathcal{I} = \left[t_0, t_\text{f}\right]$, trajectories $\bmgamma: \mathcal{I} \to \tangentbundleq: t \mapsto \left(\bmq(t), \dot{\bmq}(t)\right)^\trsp$ of the Lagrangian system are obtained by solving the \ac{ivp}
\begin{equation}
    \label{eq:lagrangianIVP}
 \begin{cases}
     \ddt\bmgamma\big\vert_{t} = 
     \bm{X}\big\vert_{\bmgamma(t)} = 
     \left(\begin{smallmatrix}
         \dot{\bmq}(t) \\\invmassmatrix\left(\bmtau - \coriolisterm - \gravityterm \right)
     \end{smallmatrix} \right)&\in \mathcal{T}_{\bmgamma(t)}\tangentbundleq, 
     \\[9pt]
    \bmgamma(t_0) = \left( \begin{smallmatrix}
        \bmq_0 \\ \dot{\bmq}_0
    \end{smallmatrix}\right) &\in \mathcal{T}\mathcalq,
 \end{cases}  
\end{equation}
with $\bm{X}\big\vert_{\bmgamma(t)}$ the vector field defined by the Euler-Lagrange equations.
Note that, for Lagrangian systems, trajectories $\bmgamma$ are lifted curves on the tangent bundle $\mathcalt\mathcalq$.  

\vspace{-0.1cm}
\subsection{Model Order Reduction}
\label{subsec:modelorderreduction}
\vspace{-0.1cm}
\ac{mor} methods consider a high-dimensional dynamical system, i.e., a \ac{fom}, with known dynamic parameters, which generates smooth trajectories $\bmgamma: \mathcal{I} \to \mathcalm$ 
characterized by an \ac{ivp} of the form
\begin{equation}
\label{eq:vectorfieldIVP}
 \begin{cases}
     \ddt\bmgamma\big\vert_{t} &= \bm{X}\big\vert_{\bmgamma(t)} \in \mathcalt_{\bmgamma(t)}\mathcalm, \quad t\in\mathcal{I}, \\[6pt]
    \bmgamma(t_0) &= \bmgamma_0 \in \mathcalm,
 \end{cases}    
\end{equation}
where $\bm{X}\big\vert_{\bmgamma(t)}$ is a smooth vector field defining the evolution of the system, so that $\bmgamma(t) \in \mathcal M$. Note that we have $\mathcalm=\mathcal T \mathcal Q$ in~\eqref{eq:lagrangianIVP}. The goal of \ac{mor} is to accurately and efficiently approximate the set of solutions
$ S = \left\{\bmgamma(t) \in \mathcalm \:\vert\: t \in \mathcal{I} \right\}\subseteq\mathcalm$
of the IVP~\eqref{eq:vectorfieldIVP}.
To this end, \ac{mor} methods learn a reduced manifold $\checkmathcalm$ with $\text{dim}(\checkmathcalm) \!=\! d \ll \text{dim}(\mathcalm) \!=\! n$ and a \ac{rom} 
$\ddt\check{\bmgamma}\big\vert_{t} = \check{\bm{X}}\big\vert_{\check{\bmgamma}(t)} \in \mathcalt_{\check{\bmgamma}(t)}\check{\mathcalm}$. 

Following~\citet{Buchfink2024MOR}, we identify $\checkmathcalm$ with an embedded submanifold $\varphi(\checkmathcalm) \subseteq \mathcalm$ via a smooth embedding ${\varphi: \checkmathcalm \to \mathcalm}$. The reduced initial value $\check{\bmgamma}_0 \!=\! \rho(\bmgamma_0)$ and vector field ${\check{\bm{X}}\big\vert_{\check{\bmgamma}(t)} = d\rho\big\vert_{\bmgamma(t)} \bm{X}\big\vert_{\bmgamma(t)}}$ are defined via the point and tangent reduction maps $\rho: \mathcalm \to \checkmathcalm$ and $d\rho\vert_{\bmx}: \mathcal{T}_{\bmx}\mathcalm \to \mathcal{T}_{\rho(\bmx)}\checkmathcalm$ associated with $\varphi$, where $\bmx \in \mathcal M$. The reduction maps must satisfy the projection properties
\begin{equation}
\label{eq:projectionproperties}
    \rho \circ \varphi = \identity{\checkmathcalm} \quad \text{ and } \quad
    d\rho\vert_{\varphi(\checkbmx)} \circ d\varphi\vert_{\checkbmx} = \identity{\mathcal{T}_{\checkbmx}\checkmathcalm}, \quad \forall \checkbmx \in \checkmathcalm.
\end{equation}
Trajectories of the original system are then obtained via the approximation $\bmgamma(t)\approx\varphi(\check{\bmgamma}(t))$.
In this paper, we consider the case where the high-dimensional dynamics are unknown. We leverage structure-preserving \ac{mor} methods to learn a non-linear reduced representation $\checkmathcalq$ of the high-dimensional Lagrangian configuration space $\mathcalq$ along with a Lagrangian ROM $(\checkmathcalq,\check{\lagrangian})$. 

A key aspect of \ac{mor} is the construction of the embedding $\varphi$ and corresponding point reduction $\rho$. \acp{ae} are well suited for representing these nonlinear mappings due to their expressivity and scalability~\citep{LeeCarlberg2020MOR,Otto2023MOR}.
An \ac{ae} consists of an encoder network ${\rho: \mathbbr{n} \to \mathbbr{d}}$ that maps a high-dimensional vector $\bm{x}\in\mathbbr{n}$ to a latent representation $\bm{z}\in\mathbbr{d}$ with $d \ll n$, and a decoder network $\varphi: \mathbbr{d} \to \mathbbr{n}$ that reconstructs an approximation of the original data. Given a dataset $\{\bm{x}_i\}_{i=1}^N$, the \ac{ae} parameters $\bm{\Psi}$ are trained by minimizing the reconstruction error
\begin{equation}
    \label{eq:AE_MSE}
    \ell_{\text{AE}}(\bm{\Psi}) = \frac{1}{N} \sum_{i=1}^N \| \varphi \circ \rho(\bm{x}_i) - \bm{x}_i\|^2.
\end{equation}
Notice that the projection properties~\eqref{eq:projectionproperties} hold approximately when the loss~\eqref{eq:AE_MSE} is small. However, in this paper, we leverage a constrained \ac{ae}~\citep{Otto2023MOR} that guarantees~\eqref{eq:projectionproperties}, which is essential when learning reduced representations of high-dimensional Lagrangian systems.

\vspace{-0.1cm}
\subsection{Lagrangian Neural Networks}
\label{subsec:lnn}
\vspace{-0.1cm}
In contrast to \ac{mor}, \ac{lnn} consider low-dimensional systems with unknown dynamics that we aim to learn. 
The key idea of \acp{lnn} is to include physics-based inductive bias into deep networks, ensuring the learned dynamics models conserve energy and lead to physically-plausible trajectories. In this context, DeLaN~\citep{Lutter19:DeLan,Lutter2023DeLaN} proposed to model the kinetic energy (via the mass-inertia matrix) and the potential energy of a Lagrangian function $\lagrangian$ as two networks $\bmM(\bmq; \bm{\theta}_{\text{T}})$ and $V(\bmq; \bm{\theta}_{\text{V}})$ with parameters $\bm{\theta}=\{\bm{\theta}_{\text{T}}, \bm{\theta}_{\text{V}}\}$. Trajectories of the learned dynamical system are then obtained via the equations of motion~\eqref{eq:equationsofmotion} by solving the \ac{ivp}~\eqref{eq:lagrangianIVP}.  
Given a set of $N$ observations $\left\{\bmq_i, \dbmq_i, \ddbmq_i, \bm{\tau}_i\right\}_{i=1}^{N}$, the networks are trained to minimize the loss
\begin{equation}
    \label{eq:LNN_MSE}
    \ell_{\text{LNN}}(\bm{\theta}) = \frac{1}{N} \sum_{i=1}^N \|\bm{f}(\bmq_i, \dbmq_i, \bmtau_i; \bm{\theta}) - \ddbmq_i\|^2 +\lambda||\bm{\theta}||_2^2,
\end{equation} 
where
${\bm{f}(\bmq, \dbmq, \bmtau; \bm{\theta})=\bmM^{-1}(\bm{q}; \bm{\theta}_{\text{T}})(\bmtau - \bm{c}(\bmq, \dbmq) - \bm{g}(\bmq; \bm{\theta}_{\text{V}}))}$ and $\lambda$ is a regularization constant.
Following this approach, the total energy  
$\mathcal{E} = \kineticenergy + \potentialenergy = \frac{1}{2}\dbmq^{\intercal} \massmatrix \dbmq + \potentialenergy,
$
of unforced systems is conserved.
Next, we propose an enhanced geometric version of DeLaN, which we then use in Section~\ref{sec:method} to learn the dynamics of a Lagrangian \ac{rom}.

%% file: texfiles/03method.tex
\vspace{-0.2cm}
\section{Geometric Lagrangian Neural Networks}
\label{sec:SPD_LNNs}
\vspace{-0.2cm}
The mass-inertia matrix $\bmM(\bmq)$ acts as the Riemannian metric on the configuration manifold $\mathcal{Q}$ for Lagrangian systems with quadratic kinetic energy structure. As such, $\bmM(\bmq)$ belongs to the manifold of \ac{spd} matrices $\SPD$ ~\citep{bhatiaSPDbook, pennecAIM}.
Existing \acp{lnn} enforce the symmetric positive-definiteness by predicting the Cholesky decomposition $\bm{L}(\bmq)$ with a Euclidean network and describing the mass-inertia matrix as $\bmM = \bm{L}\bm{L}^\trsp$. 
However, this parametrization overlooks the Riemannian geometry of both Cholesky and \ac{spd} spaces (see also App.~\ref{appendix:cholesky_vs_spd}). 
Here, we propose to learn the mass-inertia matrix via \ac{spd} networks, whose building blocks are formulated based on the Riemannian geometry of the \ac{spd} manifold.

\vspace{-0.1cm}
\subsection{Learning Positive-Definite Mass-Inertia Matrices on the SPD Manifold}
\label{subsec:learningSPDmassinertia}
\vspace{-0.2cm}
Our goal is to learn a mapping $\bmq \mapsto \bmM(\bmq): \mathcalq \to \SPD$, where $\SPD = \left\{\bm{\Sigma} \in \text{Sym}^n\:\rvert\:\bm{\Sigma} \succ \mathbf{0}\right\}$ is the Riemannian manifold of \ac{spd} matrices and $\text{Sym}^n$ is the space of symmetric matrices (see App.~\ref{appendix:spd_matrices_background} for a background on the \ac{spd} manifold and associated Riemannian operations).
To this end, we build on recent works that generalize classical neural networks to operate directly on the \ac{spd} manifold by leveraging its gyrovectorspace structure~\citep{lopez2021gyrospd,Nguyen22gyromanifolds,nguyen2024mmn++}. However, these networks consider that both inputs and outputs belong to $\SPD$.

\begin{figure}
\centering\adjustbox{trim=1.0cm 0cm 0.6cm 0cm}{\includesvg[width=\linewidth]{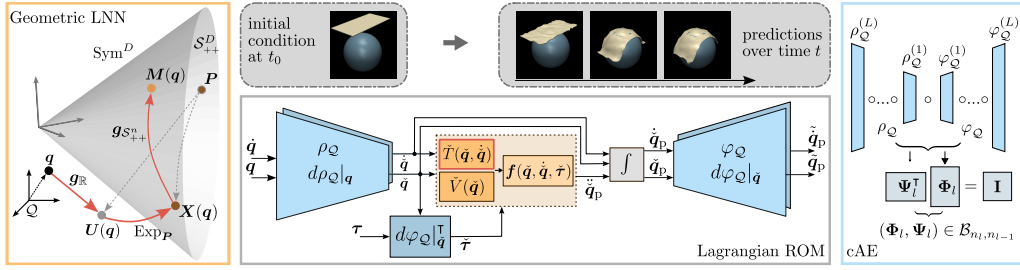}}
   \caption{Flowchart of the forward dynamics of the proposed reduced-order \ac{lnn}. The reduction mappings and embeddings of the Lagrangian \ac{rom} are depicted in blue and parametrized via a constrained \ac{ae} with biorthogonal layers (\emph{right}). The \ac{rom} dynamics are learned via a latent geometric \ac{lnn} depicted in orange. The  mass-inertia matrix is parametrized via a \ac{spd} network (\emph{left}).}
   \vspace{-0.4cm}
   \label{fig:rom_drawing}
\end{figure}

We propose a novel \ac{spd} network $\bmM(\bmq; \bm{\theta}_{\text{T}})= (g_{\SPD} \circ g_{\text{Exp}}\circ g_{\mathbb{R}}) (\bmq)$ with three consecutive components, as illustrated in Fig.~\ref{fig:rom_drawing}-\emph{left}. 
First, the input configuration $\bmq\in\mathcalq$ is fed to a standard Euclidean \ac{mlp} $g_{\mathbb{R}}:\mathbb{R}^n\to\mathbb{R}^{n(n+1)/2}$, whose output elements are identified with those of a symmetric matrix $\bm{U}\in\text{Sym}^n$. Second, $\bm{U}$ is interpreted as an element of the tangent space $\mathcal{T}_{\bm{P}}\SPD$, and mapped onto the \ac{spd} manifold via an exponential map layer $\bm{X} = \expmap{\bm{P}}{\bm{U}}$. Note that this type of layers is commonly used in hyperbolic neural networks~\citep{shimizu2021hnn++}. Third, $\bm{X}$ is fed to a set of $L_{\text{spd}}$ layers $g_{\SPD}^{(l)}:\SPD\to\SPD$, which outputs the mass-inertia matrix $\bmM\in\SPD$.  
We consider and compare two different \ac{fc} \ac{spd} layers --- which are analogous to the \ac{fc} layers in \acp{mlp} ---, namely the \ac{gyroai} layers of~\citep{lopez2021gyrospd,Nguyen22gyromanifolds}, and the \ac{gyrospd} layers of~\citep{nguyen2024mmn++}. Both can be integrated with the ReEig layers~\citep{huang2016riemannianspdnn} which are the \ac{spd} counterparts to ReLU-activations (see App.~\ref{appendix:spd_nns_layers} for details on the \ac{spd} layers).
In total, the parameters $\bm{\theta}_{\text{T}}$ of our \ac{spd} network consist of the \ac{mlp} parameters $\bm{\theta}_{\text{T}, \mathbb{R}}$, and the \ac{spd} layer parameters $\bm{\theta}_{\text{T}, \SPD} \in \SPD \times \ldots \times \SPD$, including the basepoint $\bm{P}$ of the exponential map.

\subsection{Model Training and Parameter Optimization}
\label{sec:model_training_riem_opt}
The proposed geometric \ac{lnn} models the kinetic energy via the mass-inertia matrix parametrized as a \ac{spd} network $\bmM(\bmq; \bm{\theta}_{\text{T}})$ and the potential energy via a standard Euclidean MLP $V(\bmq; \bm{\theta}_{\text{V}})$. 
The network parameters $\bm{\theta}=\{\bm{\theta}_{\text{T}}, \bm{\theta}_{\text{V}}\}$ are obtained as for classical \acp{lnn} by minimizing~\eqref{eq:LNN_MSE}.
However, as several parameters of the \ac{spd} network 
belong to $\SPD$, their geometry must be taken into account during training. To do so, we leverage Riemannian optimization~\citep{Absil07:RiemannOpt,Boumal22:RiemannOpt} to solve a problem of the form $\min_{\bm{x}\in\manifold} \ell(\bm{x})$ on a Riemannian manifold $\manifold$, where $\ell$ is the loss to minimize and $\bm{x}\in \mathcal M$ is the optimization variable. 
Conceptually, each iteration step in a first-order (stochastic) Riemannian optimization method consists of the three successive operations
\begin{equation}
\bm{\eta}_t \gets h\big(\text{grad}\: \ell(\bm{x}_t), \bm{\tau}_{t-1}\big), \quad
\bm{x}_{t+1} \gets \expmap{\bm{x}_t}{-\alpha_t \bm{\eta}_t}, \quad
\bm{\tau}_t \gets \prltrsp{\bm{x}_t}{\bm{x}_{t+1}}{\bm{\eta}_t}.
\label{eq:RiemannianGD}
\end{equation}
First, the update $\bm{\eta}_t\in\tangentspace{\bm{x}_t}$ is computed as a function $h$ of the Riemannian gradient $\text{grad}\:\ell$ at the estimate $\bm{x}_t$ and of $\bm{\tau}_{t-1}$, a parallel-transported previous update $\bm{\eta}_{t-1}\in\tangentspace{\bm{x}_{t-1}}$ to the new $\tangentspace{\bm{x}_t}$ (see App.~\ref{appendix:manifolds} for the relevant Riemannian operations). Then, an update of the estimate $\bm{x}_t$ is obtained via projecting the update $\bm{\eta}_t$, that is scaled by a learning rate learning rate $\alpha_t$ onto the manifold with the exponential map. 
Finally, the current update is parallel-transported to the tangent space of the updated estimate to prepare for the next iteration.  
Note that the function $h$ is determined by the optimization method. We use the Riemannian Adam~\citep{becigneul2018riemannianoptimization} implemented in Geoopt~\citep{geoopt} to optimize the geometric \ac{lnn} parameters. Notice that $\manifold$ is defined as a product of Euclidean and \ac{spd} manifolds to optimize the parameters $\bm{\theta}$.

\vspace{-0.2cm}
\section{Learning Reduced-order Lagrangian Dynamics}
\label{sec:method}
\vspace{-0.2cm}
Like other gray-box models, our geometric \ac{lnn} faces limitations in scaling to high-dimensional systems due to the increasing complexity of their dynamics. 
In many cases, the solutions of high-dimensional equations of motion can be well approximated by a substantially lower-dimensional surrogate dynamic model.
Building on this assumption, we propose to reduce the problem dimensionality by learning a latent dynamical system that preserves the original Lagrangian structure. 
The latent space and the reduced dynamics parameters are inferred jointly via structure-preserving \ac{mor} and a latent geometric \ac{lnn}, as explained next. The complete forward dynamics of the proposed \ac{rolnn} are illustrated in Fig.~\ref{fig:rom_drawing}. Notice that we consider physical systems with high number of \acp{dof} with observations $\{\bmq_i, \dbmq_i, \ddbmq_i,\bmtau_i\}_{i=1}^T$, and not high-dimensional representations of inherently low-dimensional state spaces as would be the case with images. 

\vspace{-0.1cm}
\subsection{Lagrangian Reduced-order Model}
\label{subsec:LagrangianMOR}
\vspace{-0.1cm}
Preserving the properties of the original \ac{fom} is crucial to ensure that the learned \ac{rom} displays similar behaviors. Therefore, we leverage the geometric framework introduced in~\cite{Buchfink2024MOR} and learn a reduced Lagrangian $(\check{\mathcalq}, \check{\lagrangian)}$ via structure-preserving \ac{mor}.
As Lagrangian trajectories are lifted curves $\bmgamma(t)=\left(\bmq(t), \dot{\bmq}(t)\right)^\trsp$, the manifold to be reduced is the tangent bundle $\tangentbundleq$. We define the lifted embedding ${\varphi: \mathcal{T}\checkmathcalq\to \mathcal{T}\mathcalq}$ for a smooth embedding $\varphiq: \checkmathcalq \to \mathcalq$ as the pair
\begin{equation}
    \label{eq:liftedembedding}
    \varphi(\checkbmq, \dcheckbmq) = (\varphiq(\checkbmq), \; d\varphiq\vert_{\checkbmq}\dcheckbmq),
\end{equation}
with $d\varphiq\vert_{\checkbmq}: \mathcalt_{\checkbmq}\checkmathcalq \to \mathcalt_{\varphi_\mathcalq(\checkbmq)}\mathcalq$ denoting the pushforward, or differential, of $\varphi_\mathcalq$ at $\checkbmq$. Analogously, we define a reduction map $\rho(\bmq, \dbmq)$ as the pair $(\rhoq(\bmq), d\rhoq\vert_{\bmq}\dbmq)$ of point and tangent reductions  $\rhoq: \mathcalq \to \checkmathcalq$ and $d\rhoq\vert_{\bmq}: \mathcal{T}_{\bmq}\mathcalq \to \mathcal{T}_{\rhoq(\bmq)}\checkmathcalq$.

Given $\varphi$ and $\rho$, the observed states $\{\bmq_i, \dbmq_i\}_{i=1}^T$ are first mapped to the \ac{rom} via the point reduction mapping to obtain reduced initial values.
Then, the reduced Lagrangian function is constructed via the pullback of the lifted embedding as
\begin{equation}
    \label{eq:reducedlagrangianpullback}
    \check{\lagrangian} = \varphi^*\lagrangian = \lagrangian \circ \varphi.
\end{equation} 
The Euler-Lagrange equations of the reduced Lagrangian yield
\begin{equation}
    \label{eq:reducedequationsofmotion}
    \check{\bmM}(\checkbmq)\ddcheckbmq + \check{\bm{c}}(\checkbmq, \dcheckbmq) + \check{\bm{g}}(\checkbmq) = \check{\bmtau}.
\end{equation}
In the intrusive case~\citep{Buchfink2024MOR}, the reduced parameters are given as a function of the known high-dimensional dynamics as $\check{\bmM}(\checkbmq) = d\varphiq\vert_{\checkbmq}^\intercal \; \massmatrix \; d\varphiq\vert_{\checkbmq}$, $\check{\bm{g}}(\checkbmq)=d\varphiq\vert_{\checkbmq}^\intercal \; \gravityterm$, $\check{\bmtau}=d\varphiq\vert_{\checkbmq}^\intercal \; \bmtau$, and $\check{\bm{c}}(\checkbmq, \dcheckbmq)$ is computed as function of $\check{\bmM}(\checkbmq)$ as in~\eqref{eq:equationsofmotion}.
Instead, we assume unknown dynamics and propose learn the reduced parameter with a geometric \ac{lnn}. Specifically, we parametrize the reduced mass-inertia matrix and the reduced potential energy of $\check{\lagrangian}$ as two networks $\check{\bmM}(\checkbmq; \bm{\theta}_{\check{\text{T}}})$ and $\check{V}(\checkbmq; \bm{\theta}_{\check{\text{V}}})$ with parameters $\bm{\theta}=\{\bm{\theta}_{\check{\text{T}}}, \bm{\theta}_{\check{\text{V}}}\}$. 
The \ac{rom}~\eqref{eq:reducedequationsofmotion} can then be used to efficiently compute reduced trajectories $\check{\bmgamma}(t)$ as solutions $\bmgamma(t) \approx \varphi(\check{\bmgamma}(t))$ of the \ac{fom}~\eqref{eq:equationsofmotion}. 
It is worth emphasizing that the preservation of the Lagrangian structure in the \ac{rom} guarantees the conservation of the reduced total energy $\check{\mathcal{E}}$ along the solutions $\check{\bmgamma}(t)$ of~\eqref{eq:reducedequationsofmotion} and the corresponding image curves $\varphi(\check{\bmgamma}(t))$ as $\check{\mathcal{E}} = \mathcal{E} \circ \varphi$. This property allows us to learn physically-consistent reduced dynamics. 
Next, we discuss how to learn $\varphiq$ and $\rhoq$. 

\vspace{-0.1cm}
\subsection{Learning the Embedding and Point Reduction}
\label{subsec:constrainedAE}
\vspace{-0.2cm}
For increased expressivity, we parametrize the point reduction $\rhoq$ and embedding $\varphiq$ of our reduced Lagrangian as the encoder and decoder of an \ac{ae}. To ensure that the reduction map $\rho(\bmq, \dbmq)$ satisfies the projection properties~\eqref{eq:projectionproperties}, we leverage the constrained AE architecture introduced by~\citet{Otto2023MOR}. The encoder and decoder networks are given as a composition of feedforward layers ${\rhoq=\rhoq^{(1)}\circ \ldots \circ \rhoq^{(L)}}$ and $\varphiq=\varphiq^{(L)}\circ \ldots \circ \varphiq^{(1)}$ with $\rhoq^{(l)}: \mathbbr{n_l} \to \mathbbr{n_{l-1}}$, $\varphiq^{(l)}: \mathbbr{n_{l-1}} \to \mathbbr{n_{l}}$, and $d = n_0 \leq \ldots \leq n_L = n$, where $d$ denotes the latent space dimension. 
Notice that the pushforward $d\varphiq\vert_{\checkbmq}$ and tangent reduction $d\rhoq\vert_{\bmq}$ are the differentials of the encoder and decoder networks. 

The key to guarantee the projection properties~\eqref{eq:projectionproperties} lies in the specific construction of layer pairs as 
\begin{equation}
    \rhoq^{(l)}(\bmq^{(l)}) = \sigma_{-}\left(\bm{\Psi}_l^{\trsp}(\bmq^{(l)} -\bm{b}_{l})\right) \quad\quad \text{and} \quad\quad
    \varphiq^{(l)}(\checkbmq^{(l-1)}) = \bm{\Phi}_{l} \sigma_{+}(\checkbmq^{(l-1)}) +\bm{b}_{l},
\end{equation}
where $(\bm{\Phi}_l, \bm{\Psi}_{l})$ and $(\sigma_{+}, \sigma_{-})$ are pairs of weight matrices and smooth activation functions, respectively, and $\bm{b}_l$ are bias vectors.
By constraining the pairs to satisfy $\bm{\Psi}_{l}^{\trsp} \bm{\Phi}_{l} = \mathbf{I}_d$ and $\sigma_{-}\circ\sigma_{+}=\identity{}$, each layer satisfies ${\rhoq^{(l)} \circ \varphiq^{(l)} = \identity{\mathbbr{n_{l-1}}}}$ and the constrained AE fullfills~\eqref{eq:projectionproperties}. 
\citet{Otto2023MOR} adhere to the first constraint by defining pairs of biorthogonal matrices $(\bm{\Phi}_{l}, \bm{\Psi}_{l})$ via an over-parametrization resulting in additional loss terms. Instead, we adopt a geometric approach that accounts for the Riemannian geometry of biorthogonal matrices. Specifically, we minimize the AE reconstruction error~\eqref{eq:AE_MSE} via Riemannian optimization (see Equation~\eqref{eq:RiemannianGD}) by considering each pair $(\bm{\Phi}_{l}, \bm{\Psi}_{l})$ as elements of the biorthogonal manifold $\mathcal{B}_{n_l,n_{l-1}}=\{ (\bm{\Phi}, \bm{\Psi})\in\mathbb{R}^{n_l \times n_{l-1}}\times\mathbb{R}^{n_l \times n_{l-1}} : \bm{\Psi}^{\trsp}\bm{\Phi}=\bm{I}_{n_{l-1}} \}$ (see App.~\ref{appendix:biorthogonal_manifold} for a background on the biorthogonal manifolds and associated operations). As it will be shown later in our experiments, optimizing the biorthogonal weights on the biorthogonal manifold is crucial when jointly optimizing the latent space and the associated reduced-order dynamic parameters.
The second constraint is met by utilizing the smooth, invertible activation functions defined in~\citep[Equation 12]{Otto2023MOR}. 
Additional details on the AE architecture, including activation functions, and layer derivatives, are provided in App.~\ref{appendix:constrained_ae}. 

\vspace{-0.1cm}
\subsection{Model Training}
\label{subsec:training}
\vspace{-0.2cm}
Finally, we propose to jointly learn the parameters $\bm{\Xi}\!=\!\{\bm{\Phi}_{l}, \bm{\Psi}_{l}, \bm{b}_{l}\}_{l=1}^{L}$ of the \ac{ae} and $\bm{\theta}\!=\!\{\bm{\theta}_{\check{\text{T}}, \mathbb{R}}, \bm{\theta}_{\check{\text{T}}, \SPDd}, \bm{\theta}_{\check{\text{V}}}\}$ of the latent geometric \ac{lnn}. We consider two losses, both of which minimize the AE reconstruction error, the latent \ac{lnn} loss, and a joint error on the reconstructed predictions.

\textbf{Training on acceleration. } Given a training set $\left\{\bmq_i, \dbmq_i, \ddbmq_i, \bm{\tau}_i\right\}_{i=1}^{N}$, the acceleration loss is
\small
\begin{equation}
    \label{eq:lossROM_acc}
       \ell_{\text{ROM, acc}}= 
        \frac{1}{N}\sum_{i=1}^N \underbrace{\|\tilde{\bmq}_i - \bmq_i \|^2 + \| \tilde{\dbmq}_i - \dbmq_i \|^2 + \|\tilde{\ddbmq}_i - \ddbmq_i\|^2}_{\ell_{\text{AE}}}
        +
        \underbrace{\|\ddcheckbmq_{\text{p},i} - \ddcheckbmq_i \|^2 }_{\ell_{\text{LNN},d}}
        + \underbrace{\|\tilde{\ddbmq}_{\text{p}, i} - \ddbmq_i\|^2}_{\ell_{\text{LNN},n}}
        + \lambda \|\bm{\theta}\|_2^2.
\end{equation}
\normalsize
$\tilde{\bmq}_i$, $\tilde{\dbmq}_i$, $\tilde{\ddbmq}_i$ are the reconstructed position, velocity, and acceleration obtained by successively applying the lifted point reduction and embedding, and its derivatives, and $\ddot{\check{\bm{q}}}_i$ is the reduced acceleration. 
The latent acceleration predictions $\ddcheckbmq_{p,i}$ are obtained from the reduced equations of motion~\eqref{eq:reducedequationsofmotion} as
\begin{equation}
\label{eq:latent_acc_pred}
    \ddcheckbmq_\text{p} = {\bm{f}(\checkbmq, \dcheckbmq, \check{\bmtau}; \bm{\theta})= \check{\bmM}^{-1}(\checkbmq; \bm{\theta}_{\check{\text{T}}})(\check{\bmtau} - \check{\bm{c}}(\checkbmq, \dcheckbmq) - \check{\bm{g}}(\checkbmq; \bm{\theta}_{\check{\text{V}}}))},
\end{equation} 
The acceleration of the \ac{fom} is then computed as $\tilde{\ddbmq}_\text{p} = d\varphiq\vert_{\checkbmq} \ddcheckbmq_\text{p} + d^2\varphiq\vert_{(\checkbmq, \dcheckbmq)} \dcheckbmq$, as shown in Fig.~\ref{fig:rom_drawing_acc}. 

\textbf{Training with multi-step integration. }
The loss~\eqref{eq:lossROM_acc} requires computing the second derivatives of the embedding and point reduction, i.e., the Hessian matrices of the AE, which result in significant computational efforts increasing with the dimension of the FOM. Moreover, it considers only single steps, while the learned dynamics are expected to predict multiple steps. Therefore, we also consider a multi-step loss that 
which numerically integrates the latent acceleration predictions~\eqref{eq:latent_acc_pred} before decoding. This is achieved via $H$ Euler forward steps with constant integration time $\Delta t$.
Given sets of observations $\left\{\bmq_i(\mathcal{I}_i), \dbmq_i(\mathcal{I}_i), \bm{\tau}_i(\mathcal{I}_i)\right\}_{i=1}^N$ over intervals $\mathcal{I}_i = \left[t_{i}, t_{i}+H\Delta t\right]$, the multi-step loss is
\vspace{-0.1cm}
\small
\begin{equation} 
    \label{eq:lossROM_ode}
    \begin{split}
        \ell_{\text{ROM,ODE}}
        = \frac{1}{HN}\sum_{i=1}^N \sum_{j=1}^H 
        &\underbrace{\|\tilde{\bmq}_{i}(t_{i,j}) - \bmq_i(t_{i,j}) \|^2 + \| \tilde{\dbmq}_{i}(t_{i,j}) - \dbmq_i(t_{i,j}) \|^2}_{\ell_{\text{AE}}}  \\
        &+  
        \underbrace{\|\dcheckbmq_{\text{p}, i}(t_{i,j}) - \dcheckbmq_i(t_{i,j}) \|^2}_{\ell_{\text{LNN},d}} + \underbrace{\|\tilde{\dbmq}_{\text{p}, i}(t_{i,j}) - \dbmq_{i}(t_{i,j})\|^2}_{\ell_{\text{LNN},n}} + \gamma\; \|\bm{\theta}\|_2^2,
    \end{split}
\end{equation}
\normalsize
\vspace{-0.1cm}
with latent velocity predictions $\dcheckbmq_{\text{p}, i}(t_{i,j}) = \int_{t_{i}}^{t_{i,j}}\bm{f}\left(\checkbmq_i(t), \dcheckbmq_i(t), \check{\bmtau}_i(t); \bm{\theta}\right)$, velocity reconstructions $\tilde{\dbmq}_{\text{p}, i}(t_{i,j})=d\varphi_\mathcal{Q}\left(\checkbmq_i(t_{i,j})\right)$, and $t_{i,j}=t_{i} + j\Delta t$. Fig.~\ref{fig:rom_drawing} illustrates the resulting forward model.

%% file: texfiles/04results.tex
\vspace{-0.2cm}
\section{Experiments}
\label{sec:experiments}
\vspace{-0.2cm}
We first evaluate the geometric \ac{lnn} on a simulated $2$-\ac{dof} planar pendulum. Second, we evaluate our geometric \ac{rolnn} to learn the dynamics of three simulated high-dimensional systems: a $16$-\ac{dof} pendulum, a $192$-\ac{dof} rope, and a $600$-\ac{dof} thin cloth. Our experiments demonstrate the ability of our approach to learn reduced-order dynamics, resulting in accurate long-term predictions. Moreover, they highlight the importance of geometry as additional inductive bias in both \ac{lnn} and \ac{ae}. Details about simulation environments, datasets, network architectures, and training are provided in App.~\ref{appendix:additional_experimental details} for each experiment. Additional results are provided in App.~\ref{app:additional_results}.
A video and source code are available at \url{https://sites.google.com/view/reduced-lagrangians}. 

\vspace{-0.2cm}
\subsection{Learning Lagrangian Dynamics with Geometric \acp{lnn}}
\label{subsec:DoublePendulum}
\vspace{-0.2cm}
We start by evaluating the long-term prediction accuracy of the geometric \ac{lnn} introduced in Section~\ref{sec:SPD_LNNs} on a $2$-\ac{dof} planar pendulum. Train and test trajectories are obtained by solving its equations of motion with randomly sampled initial positions and zero velocities. 
We compare the performance of our geometric \ac{lnn} against DeLaN, which parametrizes the mass-inertia matrix via a Cholesky network. We consider two variants, where the kinetic and potential energy networks are independent~\citep{Lutter2023DeLaN} or share parameters ($\bm{\theta}_{\text{T}} \cap \bm{\theta}_{\text{V}}$)~\citep{Lutter19:DeLan}. Moreover, we evaluate the performance of different architectures for our \ac{spd} network $\bmM(\bmq; \bm{\theta}_{\text{T}})$. First, we compare two exponential map layers, where the basepoint $\bm{P}\in\SPD$ is either set as $\mathbf{I}$ or learned as a parameter. Second, we evaluate the influence of the \ac{spd} layers by comparing performances without any \ac{spd} layers, and with \ac{gyroai}+ReEig, \ac{gyrospd}, or \ac{gyrospd}+ReEig layers.  

\begin{figure}[tbp]
	\centering
    \vspace{-0.2cm}
        \resizebox{\textwidth}{!}{
		\begin{tabular}{lllll}
            \choleskyblueline~Cholesky & \choleskyorangeline~Cholesky $\bm{\theta}_\text{T} \cap \bm{\theta}_\text{V}$& \expgreenline~$\expmapblank{\bm{I}}$ & \exppinkline~$\expmapblank{\bm{P}_{\bm{\theta}}}$ & \spdblueline~$\expmapblank{\bm{P}_{\bm{\theta}}}$,~\ac{gyrospd}
           \\ \spdredline~$\expmapblank{\bm{P}_{\bm{\theta}}}$,~\ac{gyrospd}+ReEig & \spdgreenline~$\expmapblank{\bm{P}_{\bm{\theta}}}$,~\ac{gyroai}+ReEig & \odeyellowline~$\expmapblank{\mathbf{I}},~H_\text{train}=1$ & \odepurpleline~$\expmapblank{\mathbf{I}},~H_\text{train}=8$ \\
        \end{tabular}
}

         \begin{subfigure}[b]{\textwidth}
   \centering
		\adjustbox{trim=0cm 0cm 0cm 0cm}{\includesvg[width=\linewidth]{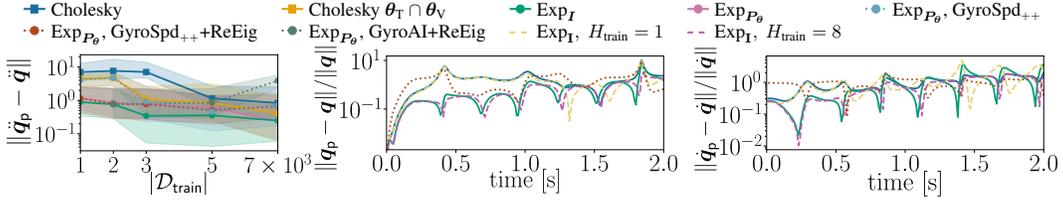}}
    \end{subfigure}
	\caption{$2$-\ac{dof} pendulum: \emph{Left}: Median acceleration prediction error for different \acp{lnn} and training set sizes $\Dtrain$ over $10$ test trajectories. Shaded regions represent first and third quartiles. 
    \emph{Middle, right:} Relative error of numerically-integrated position and velocity predictions with respect to the ground-truth trajectory over a prediction horizon $H_{\text{test}}=2000$.}
    \vspace{-0.3cm}\label{fig:spd_abl:test_error_over_set_size_unact}
\end{figure}

Fig.~\ref{fig:spd_abl:test_error_over_set_size_unact}-\emph{left} shows the acceleration prediction errors on testing data for selected architectures trained on acceleration data.
We observe that our geometric \acp{lnn} consistently outperform both DeLaNs, especially in low-data regimes.
Interestingly, the geometric \acp{lnn} with \ac{spd} layers do not noticeably outperform those employing solely Euclidean and exponential-map layers (see App.~\ref{appendix:spd_abl:on_spd_man_nns} for an extensive analysis on \ac{spd} layers and App.~\ref{appendix:spd_abl:runtimes} for training times).
Fig.~\ref{fig:spd_abl:test_error_over_set_size_unact}-\emph{middle, right} depict long-term trajectory predictions obtained via Euler integration of the predicted state derivatives. In addition to the aforementioned models, we consider several geometric \acp{lnn} trained with multi-step integration.
We observe that the geometric \acp{lnn} lead to significantly better long-term predictions than the DeLaNs, with the \ac{spd} layers following the previously-observed performance trend (see also App.~\ref{appendix:spd_abl:unactuated}). Moreover, the geometric \ac{lnn} trained over $8$ integration steps outperforms those trained on accelerations, and results in the most accurate predictions (see also App.~\ref{appendix:spd_abl:learn_from_horizon}). 
Overall, our results validate the effectiveness of considering additional inductive bias given by \emph{(1)} the intrinsic geometry of the mass-inertia matrix, and \emph{(2)} multi-step integration when training \acp{lnn}. These results are further validated on a second dataset in App.~\ref{appendix:spd_abl:sine_tracking}.

\vspace{-0.1cm}
\subsection{Learning Reduced-Order Lagrangian Dynamics}
\label{subsec:experiments_lagrangian_rom}
\vspace{-0.1cm}
Next, we learn reduced-order dynamics of several simulated high-dimensional physical systems.
Due to the limited performance improvements and increased computational complexity of the \ac{spd} layers, we only evaluate \ac{spd} networks $\bmM(\bmq; \bm{\theta}_{\text{T}})= (g_{\text{Exp}}\circ g_{\mathbb{R}}) (\bmq)$ with $L_{\text{T}, \SPD}=0$.

\vspace{-0.2cm}
\subsubsection{Coupled Pendulum ($16$ \acp{dof})}
\label{subsec:coupled_pend}
\vspace{-0.1cm}
As a first \ac{rom} example, we consider a $16$-\ac{dof} pendulum. 
We ensure that the dynamics are reducible by constraining the motion of the last $12$-\ac{dof} as nonlinear combinations of the first $4$-\ac{dof}. As such, the high-dimensional pendulum is reducible to $4$ latent dimensions.

\begin{figure}[tbp]
        \centering
        \adjustbox{trim=0cm 0cm 0cm 0cm}{
            \includesvg[width=.9\linewidth]{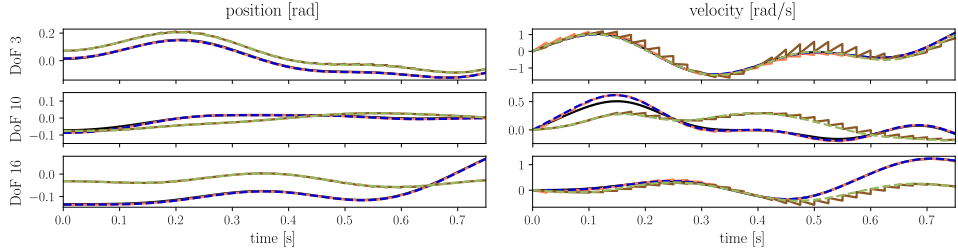}  
        }
        \caption{$16$-\ac{dof} pendulum: 
        Comparisons of the position and velocity predictions from the \acp{rolnn} trained on acceleration (\solidbrownline) and via multi-step integration (\orangeline) with the ground truth (\blackline). The corresponding \ac{ae} reconstructions (\dottedgreenline) and (\darkblueline) are depicted for completeness. The model is updated with a new initial condition $(\bmq_0, \dbmq_0)^{\trsp}$ every $\SI{0.025}{s}$ ($H_{\text{test}}=25$).}
        \vspace{-0.3cm}
\label{fig:coupled_pend:pos_k25}
\end{figure} 

\begin{table}[tbp]
\centering
\vspace{-0.2cm}
\caption{Comparison of mean and standard deviation of prediction errors over $10$ test trajectories.}
\label{tab:coupled:acc_pred_comparison_ROM_FOM}
\resizebox{\columnwidth}{!}{
\begin{tabular}{cccccc}
 & \ac{lnn} & L-OpInf & L-OpInf (LNN, ODE) & \ac{rolnn} (acc) & \ac{rolnn} (ODE) 
\\
\midrule
$\|\tilde{\ddbmq}_{\text{p}} - \ddbmq\| / \|\ddbmq\|$ & $(1.97 \pm 1.49)\times 10^{2}$ & ---  & --- & $(5.87 \pm 4.74)\times 10^{-1}$& \bm{$(3.53 \pm 2.17)\times 10^{-1}$}\\ 
$\|\tilde{\dbmq}_{\text{p}} - \dbmq\| / \|\dbmq\|$ &  --- & ---  & --- & $(1.33 \pm 0.79)\times 10^{0}$& $\bm{(3.24 \pm 3.43)\times 10^{-1}}$ \\ 
$\|\tilde{\bmq}_{\text{p}} - \bmq\| / \|\bmq\|$ & ---  & $ (1.31 \pm 0.85)\times 10^{1}$ & $(2.45 \pm 0.96) \times 10^{-1}$ & $(3.07 \pm 1.86)\times 10^{-1}$ & $\bm{(9.39 \pm 3.16)\times 10^{-3}}$\\ 
\end{tabular}
}
\end{table}

\textbf{Learning High-Dimensional Dynamics.}  
We model the high-dimensional dynamics with our \ac{rolnn}. The parameters of the structure-preserving \ac{ae} and latent geometric \ac{lnn} are jointly trained via Riemannian optimization. We set the latent space dimension to $d=4$ and the exponential map basepoint to $\mathbf{I}$ (see App.~\ref{appendix:coupled:details} for implementation details).
We consider models trained on acceleration data via~\eqref{eq:lossROM_acc}, and with multi-step integration using~\eqref{eq:lossROM_ode} with $H_{\text{train}}=8$. 
Fig.~\ref{fig:coupled_pend:pos_k25} shows the \ac{ae} reconstruction and \ac{rolnn} predictions ($H_{\text{test}}=25$) for selected \acp{dof} of a test trajectory. Average testing errors for different $H_{\text{test}}$ are reported in App.~\ref{appendix:coupled:latent_pred_horizon}. We observe that the \ac{rolnn} trained with multi-step integration consistently outperforms the \ac{rolnn} trained on accelerations, validating the importance of considering successive states during training.  
Moreover, we compare our \ac{rolnn} with full-order \acp{lnn} that directly learn high-dimensional dynamic parameters. As shown in Table~\ref{tab:coupled:acc_pred_comparison_ROM_FOM}, even the best-performing \ac{fom} produces prediction errors that are orders of magnitude higher than the \ac{rolnn} errors. Such erroneous acceleration predictions did not allow us to obtain stable velocity and position predictions.
We also compare our models with the Lagrangian operator inference (L-OpInf)~\citep{sharma24}, and a variant of L-OpInf that uses the same linear projection method with a latent LNN for parameter identification~\citep{sharma24NN}. For fair comparisons, we replace the latent network used in~\citep{sharma24NN} by a geometric LNN trained with multi-step integration (see App.~\ref{appendix:coupled:architecture}). 
L-OpInf shows significantly higher errors than the RO-LNNs. While L-OpInf with LNN reaches lower errors, they still remain higher than for the similarly-trained RO-LNN (ODE), thus showcasing the limitations of linear subspaces.

\textbf{Joint training.} 
It is important to emphasize that the quality of the \ac{ae}-reconstructed states is crucial for learning accurate dynamics. If the \ac{rom} cannot effectively capture the solution space of the \ac{fom}, the learned dynamics may systematically
deviate from the ground truth. 
Here, we analyze the influence of the \ac{ae} on the overall performance of our \ac{rolnn}. Specifically, we compare our model with the joint training process with \emph{(1)} a \ac{rom} which sequentially trains the independent constrained \ac{ae} and latent \ac{lnn}, and \emph{(2)} a jointly-trained \ac{rom} that employs the overparametrization of~\citep{Otto2023MOR}, and thus overlooks the geometry of the biorthogonal \ac{ae} weights.
All models are trained with $d=4$ via multi-step integration with $H_{\text{train}}=8$ and $\Delta t =10^{-3}\SI{}{\s}$.

\begin{figure}[tbp]
    \centering
        \centering
        \adjustbox{trim=0cm 0cm 0cm 0cm}{
            \includesvg[width=.9\linewidth]{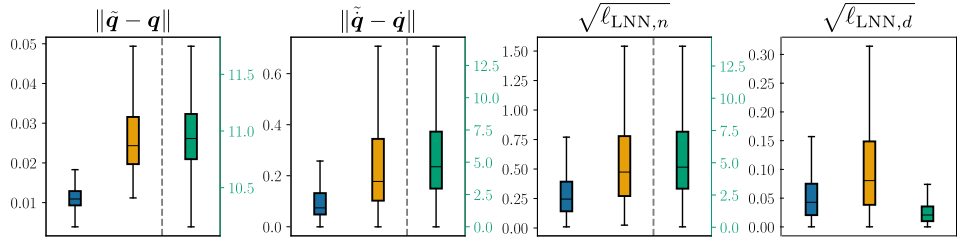}
        }
    \caption{Median and quartiles of the errors of jointly-trained (on-biorthogonal-manifold (\bluebox) vs overparametrized (\orangebox)) and sequentially-trained (\greenbox) models. Note that the latter have their own $y$-axis in the left three plots, which is at least an order of magnitude higher.}
    \vspace{-0.2cm}
    \label{fig:coupled_pend:ae_abl_vel}
\end{figure}

Fig.~\ref{fig:coupled_pend:ae_abl_vel} shows \ac{ae} reconstruction errors (position and velocity) along with the high-dimensional and latent prediction errors averaged over $10$ testing trajectories. We observe that the jointly-trained models outperform the sequentially-trained one. Although the latter learns accurate dynamics in the latent space, the reconstructed high-dimensional dynamics result in significantly higher prediction errors (see the right $y$-scale). 
Moreover, the overparametrized model leads to higher errors in all loss components compared to the \ac{rolnn} trained on the biorthogonal manifold. The overparametrized \ac{ae} is also noticeably harder to train jointly with the latent \ac{lnn}: In our experiments, $4$ out of $6$ trainings led to nan-values. Altogether, our results suggest that considering the intrinsic geometry of the biorthogonal manifold enhances the performance and stability of the parameter optimization (see App.~\ref{appendix:coupled:ae_pos_rec_comparison} for further comparisons).

\vspace{-0.2cm}
\subsubsection{Rope ($192$ \acp{dof})}
\label{subsec:rope}
\begin{figure}[tbp]
    \centering
    \includegraphics[width=\linewidth]{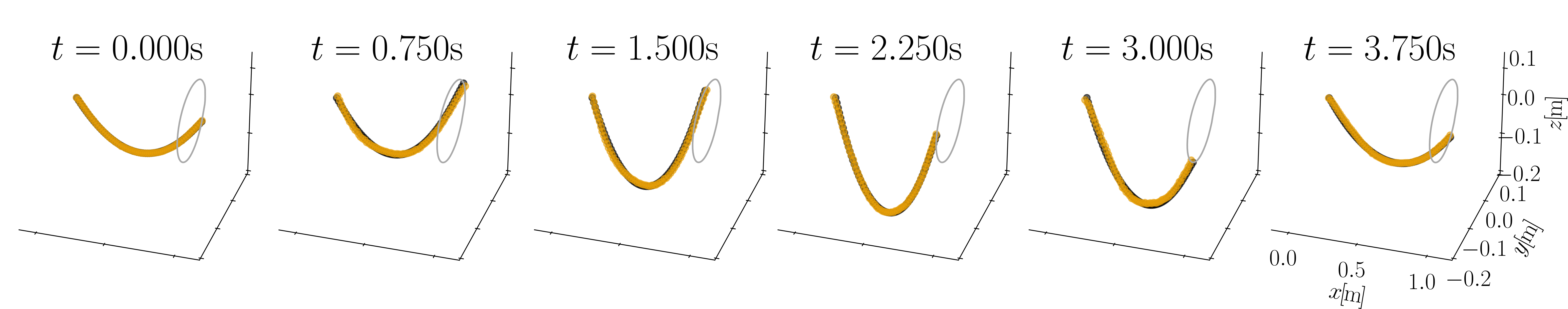}
    \caption{Predicted rope position (\orangeline) and ground truth (\blackline) at selected timesteps for a prediction horizon $H_{\text{test}}=25$. 
    The grey circle depicts the circular trajectory of the end of the rope.}
    \vspace{-0.3cm}
    \label{fig:rope:frames_visualization}
\end{figure}

\vspace{-0.1cm}
Next, we consider high-dimensional deformable systems and learn the dynamics of a $192$-dimensional rope. Trajectories are generated by tracing a circle with one end while keeping the other fixed. We train our \ac{rolnn} with $d=10$ via multi-step integration (see App.~\ref{appendix:rope} for details on the data and implementation). 
Fig~\ref{fig:rope:frames_visualization} shows the predicted rope configurations for a horizon of $\SI{0.025}{s}$ ($H_{\text{test}}=25$). Our model accurately predicts the high-dimensional dynamics of the rope (see also the predictions for selected \acp{dof} in App.~\ref{appendix:rope:results}).
Notice that our attempts to train a full-order \ac{lnn} were hindered by the high dimensionality of the system. Similarly, our attempts to train L-OpInf did not result in any stable solution, thus substantiating the limitations of linear methods.

\textbf{Latent space dimension.} 
Selecting an appropriate latent dimension is an important consideration for \ac{mor}.
Here, we study the influence of the latent dimension on the performance of our \ac{rolnn}. 
Table~\ref{tab:rope:loss_components_over_latent_dim} reports average reconstruction and prediction errors of \acp{rolnn} with latent dimensions $d=\{4,6,10,14\}$ on $10$ testing trajectories.
We observe that errors initially decrease as the latent dimension increases, suggesting that higher-dimensional latent spaces better capture the original high-dimensional dynamics. However, the errors increase beyond a certain latent dimension, indicating that the latent \ac{lnn} becomes harder to train. In other words, the choice of latent dimension trades off between latent space expressivity and the limitations of \acp{lnn} in higher dimensions.

\begin{table}[tbp]
\centering
\caption{Rope reconstruction errors and prediction loss components from \eqref{eq:lossROM_ode} for \acp{rolnn}.}
\label{tab:rope:loss_components_over_latent_dim}
\resizebox{\columnwidth}{!}{
\begin{tabular}{ccccc}
\toprule
$d$& $\|\tilde{\bmq} - \bmq \|^2$ & $\|\tilde{\dbmq} - \dbmq \|^2$ & $\ell_{\text{LNN},n}$ & $\ell_{\text{LNN},d}$ \\
\midrule
$4$ & $2.95\times 10^{-2} \pm 3.24\times 10^{-2}$ & $4.63\times 10^{-2} \pm 7.69\times 10^{-2}$ &$5.06^{-2} \pm 8.23\times 10^{-2}$ & $3.00\times 10^{-3} \pm 1.34\times 10^{-2}$\\[0.2mm]
$6$  & $8.19\times 10^{-3} \pm 9.11\times 10^{-3}$ & $1.52\times 10^{-2} \pm 2.92\times 10^{-2}$ & $2.40\times 10^{-2} \pm 4.57\times 10^{-2}$ & $\bm{1.81\times 10^{-3} \pm 5.69\times 10^{-3}}$\\[0.2mm]
$10$  & $\bm{4.75\times 10^{-3} \pm 5.73\times 10^{-3}}$ & $\bm{6.53\times 10^{-3} \pm 1.62\times 10^{-2}}$ & $\bm{1.43\times 10^{-2} \pm 3.45\times 10^{-2}}$ & $2.87\times 10^{-3} \pm 8.92\times 10^{-3}$\\ [0.2mm]
$14$ & $6.16\times 10^{-2} \pm 6.52\times 10^{-2}$ & $ 5.66\times 10^{-2}\pm 8.48\times 10^{-2}$ & $9.22\times 10^{-1} \pm 1.19\times 10^{-1}$ & $5.06\times 10^{-2} \pm 1.08\times 10^{-1}$ \\
\bottomrule
\end{tabular}
}
\vspace{-0.2cm}
\end{table}

\vspace{-0.2cm}
\subsubsection{Cloth ($600$ \acp{dof})}
\label{subsec:cloth}
\begin{figure}[t]
    \centering
    \adjustbox{trim=0cm 0cm 0cm 0cm}{\includesvg[width=0.75\textwidth]{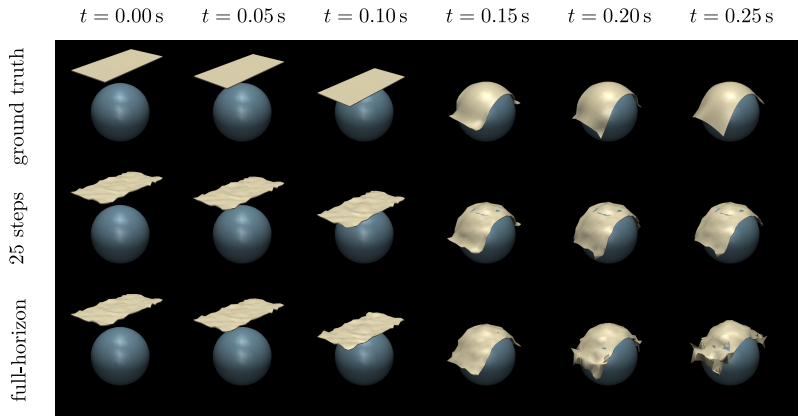}}
    \caption{Predicted cloth configuration at selected times for $25$-steps and $2500$-steps horizons.}
    \label{fig:cloth:mj:frames}
    \vspace{-0.3cm}
\end{figure}

\vspace{-0.1cm}
As a final example, we learn the dynamics of an even higher-dimensional deformable system: A simulated $600$-\ac{dof} thin cloth falling onto a sphere. We generate different trajectories by randomly varying the sphere radius, which lead to different actuations applied to the deformable structure. We train our \ac{rolnn} with $d=10$ via multi-step integration on a set of $|\mathcal{D_{\text{train}}}| = 8000$ samples (see App.~\ref{appendix:cloth} for details on data and implementation). 
Fig.~\ref{fig:cloth:mj:frames}-\emph{middle} depicts the predicted cloth configurations for a horizon of $\SI{0.0025}{s}$ ($H_{\text{test}}=25$), showing that our model accurately predicts the cloth high-dimensional dynamics (see also the predictions for selected \acp{dof} and errors in App.~\ref{appendix:cloth:results}).

\textbf{Full-horizon predictions.} 
We evaluate the prediction quality of our model over a longer time horizon. Namely, given an initial state $(\bmq_0, \dbmq_0)^{\trsp}$, we use our model to predict the motion of the cloth for $H_{\text{test}}=2500$ timesteps. 
Fig.~\ref{fig:cloth:mj:frames}-\emph{bottom} shows the cloth configurations predicted over the full horizon (see App.~\ref{appendix:cloth:results} for predictions of selected \acp{dof} and energy).
Our model accurately predicts the cloth high-dimensional dynamics over the full horizon, suggesting that it successfully learned the constraints and actuation effects of the cloth falling on a sphere.

%% file: texfiles/05conclusion.tex
\vspace{-0.2cm}
\section{Discussion} 
\label{sec:conclusion}
\vspace{-0.2cm}
This paper proposed a novel geometric network architecture for learning the dynamics of high-dimensional Lagrangian systems. 
By leveraging geometry and physics as inductive bias, our model is physically-consistent and conserves the reduced energy, infers interpretable reduced dynamic parameters, and effectively learns reduced-order dynamics that accurately describe the behavior of high-dimensional rigid and deformable systems. 
It is worth emphasizing the ubiquitous role of geometry in our model. First, considering the geometry of Lagrangian systems allows us to define nonlinear embeddings that ensure the preservation of the energy along solution trajectories and the interpretability of the model. Second, considering the geometry of full-order \acp{lnn}, latent \acp{lnn}, and \ac{ae} parameters during their (joint) optimization leads to increased performance, especially at low-data regimes, in addition to guaranteeing that they belong to the manifold of interest.

It is important to note that the prediction quality can only be as good as the \ac{ae} reconstruction. In other words, 
if the \ac{rom} cannot effectively capture the \ac{fom} solution space, the learned dynamics will systematically deviate from the ground truth.
In this sense, the selection of the latent dimension is crucial, but non-trivial due to the unknown degree of coupling in the \ac{fom}.
In future work, we plan to extend our approach to applications beyond system identification. In particular, we will work on the design control strategy based on the learned reduced dynamics for robotic manipulation of deformable objects and dynamic control of soft robots.

%% file: texfiles/technical_appendix.tex
\section{Riemannian Manifolds}
\label{appendix:manifolds}
\subsection{The Manifold of SPD Matrices}
\label{appendix:spd_matrices_background}
In this section, we provide a brief overview on the Riemannian manifold of symmetric positive-definite (SPD) matrices. 
The set of \ac{spd} matrices is defined as $\SPD = \left\{\bm{\Sigma} \in \text{Sym}^n \:\rvert\:\bm{\Sigma} \succ \mathbf{0}\right\}$ with $\text{Sym}^n$ denotes the set of symmetric $\mathbbr{n \times n}$ matrices. The set $\SPD$ forms a smooth manifold of dimension $\text{dim}(\SPD) = \frac{n(n+1)}{2}$, which can be represented as the interior of a convex cone embedded in $\text{Sym}^n$. 
The tangent space $\mathcal{T}_{\bm{\Sigma}}\SPD$ at any point ${\bm{\Sigma}\in\mathcal{S}_{\ty{++}}^n}$ is identified with the symmetric space $\text{Sym}^n$ with origin at $\bm{\Sigma}$.

The \ac{spd} manifold can be endowed with various Riemannian metrics. In this paper, we consider the affine-invariant metric~\citep{pennecAIM}, which is widely used due to its excellent theoretical properties. The affine-invariant metric defines the inner product $g:\mathcal{T}_{\bm{\Sigma}} \SPD \times \mathcal{T}_{\bm{\Sigma}} \SPD \to \mathbb{R}$ given, for two matrices $\bm{T}_1$, $\bm{T}_2\in\mathcal{T}_{\bm{\Sigma}} \SPD$, as
\begin{equation}
\langle \bm{T}_1,\bm{T}_2 \rangle_{\bm{\Sigma}} \;\;=\;\; \tr(\bm{\Sigma}^{-\frac{1}{2}}\bm{T}_1\bm{\Sigma}^{-1}\bm{T}_2\bm{\Sigma}^{-\frac{1}{2}}).
\label{Eq:SPDinnerprod}
\end{equation} 

The affine-invariant metric defines the following geodesic distance
\begin{equation}
\manifolddist{\bm{\Lambda}}{\bm{\Sigma}} = \|\log(\bm{\Sigma}^{-\frac{1}{2}}\bm{\Lambda}\bm{\Sigma}^{-\frac{1}{2}})\|_\text{F}.
\label{Eq:SPDdist}
\end{equation}
To map back and forth between the Euclidean tangent space and the manifold, we employ the corresponding exponential and logarithmic maps $\expmapblank{\bm{\Sigma}}: \mathcal{T}_{\bm{\Sigma}} \SPD\to \SPD$ and $\logmapblank{\bm{\Sigma}}: \SPD\to \mathcal{T}_{\bm{\Sigma}}\SPD$. 
The exponential and logarithmic maps are then computed in closed form as
\begin{align}
\bm{\Lambda} & = \expmap{\bm{\Sigma}}{\bm{S}} = \bm{\Sigma}^{\frac{1}{2}}\exp(\bm{\Sigma}^{-\frac{1}{2}}\bm{S}\bm{\Sigma}^{-\frac{1}{2}})\bm{\Sigma}^{\frac{1}{2}}, \\
\bm{S} & = \logmap{\bm{\Sigma}}{\bm{\Lambda}} = \bm{\Sigma}^{\frac{1}{2}}\log(\bm{\Sigma}^{-\frac{1}{2}}\bm{\Lambda}\bm{\Sigma}^{-\frac{1}{2}})\bm{\Sigma}^{\frac{1}{2}},
\label{Eq:SPDmaps}
\end{align}
where $\exp(\cdot)$ and $\log(\cdot)$ denote the matrix exponential and logarithm functions.
The parallel transport $\prltrspblank{\bm{\Sigma}}{\bm{\Lambda}}: \mathcal{T}_{\bm{\Sigma}}\SPD\to\mathcal{T}_{\bm{\Lambda}}\SPD$ allowing us to move elements between different tangent spaces is defined as 
\begin{equation}
\bm{\tilde{T}} = \prltrsp{\bm{\Sigma}}{\bm{\Lambda}}{\bm{T}} = \bm{A}_{\bm{\Sigma}\to\bm{\Lambda}} \; \bm{T} \; \bm{A}_{\bm{\Sigma}\to\bm{\Lambda}}^\trsp,
\label{Eq:spdPT}
\end{equation}
with $\bm{A}_{\bm{\Sigma}\to\bm{\Lambda}} = \bm{\Lambda}^{\frac{1}{2}}\bm{\Sigma}^{-\frac{1}{2}}$.
The exponential map, logarithmic map, and parallel transport are key operations for optimizing parameters on the \ac{spd} manifold in Sections~\ref{sec:SPD_LNNs} and~\ref{sec:method}.

\subsection{The Biorthogonal Manifold}
\label{appendix:biorthogonal_manifold}
Pairs of matrices matrices $\bm{\Phi}, \bm{\Psi}\in\mathbbr{n\times d}$ with $n\geq d \geq 1$ that fullfil $\bm{\Psi}^\trsp\bm{\Phi} = \mathbf{I}$ form the smooth manifold $\mathcal{B}_{n,d}=\{ (\bm{\Phi}, \bm{\Psi})\in\mathbb{R}^{n \times d}\times\mathbb{R}^{n \times d}\:\rvert\:\bm{\Psi}^{\trsp}\bm{\Phi}=\bm{I}_d \}$~\citep{Otto2023MOR}. The biorthogonal manifold $\mathcal{B}_{n,d}$ is an embedded submanifold of the Euclidean product space $\mathbbr{n\times d} \times \mathbbr{n\times d}$ of dimension $\text{dim}\left(\mathcal{B}_{n,d}\right) = 2nd - d^2$.
The tangent space at a point $\biopair \in \bio$ is given by
\begin{equation}
    \label{eq:bio_tangentspace}
    \Tbio = \left\{\left(\bm{V}, \bm{W}\right) \in \mathbbr{n\times d}\times \mathbbr{n \times d}\;:\; \bm{W}^\trsp\bm{\Phi} + \bm{\Psi}^\trsp\bm{V} = \mathbf{0}\right\}.
\end{equation}

Vector can be projected from the embedding space onto the tangent space via the projection operation $\text{Proj}_{\biopair} :\mathbb{R}^{n \times d}\times\mathbb{R}^{n \times d} \to \Tbio$ is given by 
\begin{equation}
    \label{eq:bio_man_proj}
    \projblank{\biopair}{\bm{X},\bm{Y}} = \left(\bm{X} - \bm{\Psi}\bm{A}, \bm{Y} - \bm{\Phi}\bm{A}^\trsp\right),
\end{equation}
where $\bm{A}$ is the solution of the Sylvester equation
$\bm{A}(\bm{\Phi}^\trsp\bm{\Phi}) + (\bm{\Psi}^\trsp\bm{\Psi})\bm{A} = \bm{Y}^\trsp \bm{\Phi} + \bm{\Psi}^\trsp\bm{X}$.

Riemannian operations on the biorthogonal manifold are generally difficult to obtain. Therefore, we leverage a retraction $\mathrm{R}_{\biopair}:\Tbio \to \bio$ as first-order approximation for the exponential map. The retraction is formulated as
\begin{equation}
    \label{eq:retraction_bio}
    \retrblank{\biopair}{\bm{V}, \bm{W}} = \left((\bm{\Phi} + \bm{V})\left[(\bm{\Psi + \bm{W}})^\trsp(\bm{\Phi}+\bm{V})\right]^{-1}, \bm{\Psi} + \bm{W}\right).
\end{equation}
A first-order approximation of the parallel transport operation on the biorthogonal manifold $\prltrspblank{\left(\bm{\Phi}_1, \bm{\Psi}_1\right)}{\left(\bm{\Phi}_2, \bm{\Psi}_2\right)}: \mathcal{T}_{\left(\bm{\Phi}_1, \bm{\Psi}_1\right)}\bio \to \mathcal{T}_{\left(\bm{\Phi}_2, \bm{\Psi}_2\right)}\bio$ is then defined via the successive application of the retraction and projection as
$\prltrspblank{\left(\bm{\Phi}_1, \bm{\Psi}_1\right)}{\left(\bm{\Phi}_2, \bm{\Psi}_2\right)} = \text{Proj}_{\left(\bm{\Phi}_2, \bm{\Psi}_2\right)} \circ \mathrm{R}_{\left(\bm{\Phi}_1, \bm{\Psi}_1\right)}$.

\section{Drawbacks of the Cholesky Decomposition in \acp{lnn}}
\label{appendix:cholesky_vs_spd}
The Cholesky decomposition $\bm{M} = \bm{L} \bm{L}^\trsp $ with lower-triangular matrix $\bm L\in\mathbb{R}^{n\times n}$ is commonly-employed to parametrize symmetric positive-definite matrices $\bm{M} \in \SPD$. For instance, DeLaN~\cite{Lutter19:DeLan,Lutter2023DeLaN} enforces the symmetric positive-definiteness of the mass-inertia matrix $\bmM$ by inferring the lower-triangular matrix via a Cholesky network $\bm{L}(\bmq;\theta_{T})$ and reconstructing the mass-inertia matrix as $\bmM=\bm{L} \bm{L}^\trsp$.  While inferring $\bm{L}$ guarantees the symmetric positive-definiteness of $\bmM$, this method defines the distances between two SPD matrices based on the Euclidean distance between their Cholesky decompositions, therefore neglecting the curvature of $\SPD$. This choice of distance suffers from the problematic swelling effect~\citep{feragenfuster2017, linSPD2019}, where the volume of the SPD matrices grows significantly while interpolating between two SPD matrices of identical volumes. Solutions inferred via a Cholesky decomposition will suffer from this swelling effect and lead to inaccurate predictions of the dynamics. These undesired effect can be avoided by directly inferring the mass-inertia matrix $\bmM$ in the SPD manifold $\SPD$ equipped with the affine-invariant metric~\citep{pennecAIM}. To do so, we parametrize the mass-inertia matrix with a SPD network $\bmM(\bmq;\theta_T)$, as explained in Section~\ref{subsec:learningSPDmassinertia}, and by optimizing its parameters using Riemannian optimization methods, as explained in Section~\ref{sec:model_training_riem_opt}.

\section{Layer types in SPD-Networks}
\label{appendix:spd_nns_layers}
Here, we briefly introduce the different layers used in the SPD networks of Section~\ref{sec:SPD_LNNs}.

\textbf{Euclidean Layers.} Our SPD network leverages classical fully-connected layers to model functions that return elements on the tangent space of a manifold. The output of the $l$-th Euclidean layer $\bm{x}^{(l)}$ is given by 
\begin{equation}
    \label{eq:ff_nn_euclideanlayer}
    \bm{x}^{(l)} = \sigma(\bm{A}_{l}\bm{x}^{(l-1)} + \bm{b}_{l}),
\end{equation}
with $\bm{A}_{l} \in\mathbbr{n_{l}\times n_{l-1}}$ and $\bm{b}_{l}\in\mathbbr{n_{(l)}}$ the weight matrix and bias of the layer $l$, and $\sigma$ a nonlinear activation function of choice. 

\textbf{Exponential Map Layers.} In the case of \ac{spd} manifolds, the exponential map layer is used to map layer inputs $\bm{X}^{(l-1)} \in \text{Sym}^n$ from the tangent space to the manifold. The layer output is given by
\begin{equation}
    \label{eq:exp_layer}
    \bm{X}^{(l)} = \text{Exp}_{\bm{P}}(\bm{X}^{(l-1)}),
\end{equation}
with $\bm{P}\in \SPD$ denoting the basepoint of the considered tangent space. One option is to define $\bm{P}$ as equal to the identity matrix $\mathbf{I}$. In that case, the layer input is assumed to lie in the tangent space at the origin of the cone. As the tangent spaces only give a local approximation to the manifold's curvature, the basepoint can instead be learned as a network parameter $\bm{P} \in \bm{\theta}_{\text{T, $\SPD$}}$ to potentially increase expressivity.

\textbf{Gyrocalculus-based \ac{fc} \ac{spd} Layers (\ac{gyroai}).} Together with gyrocalculus formulations for the SPD manifold, \cite{lopez2021gyrospd} and \cite{Nguyen22gyromanifolds} presented a version of fully connected layers for SPD data. With $\oplus$ as addition and $\otimes$ as scaling operations for SPD gyrovectors,
\begin{equation}
    \label{eq:gyronl_layer}
    \bm{X}^{(l)} = \left(\bm{A}_{l}\otimes \bm{X}^{(l-1)}\right) \oplus \bm{B}_{l},
\end{equation}
defines the $l$-th gyrolayer output $\bm{X}^{(l)}\in \SPDn$ for the input $\bm{X}^{(l-1)}\in \SPDn$. Here,
the gyrovector operations correspond to $\bm{A}\otimes \bm{X} = \text{exp}(\bm{A}*\text{log}(\bm{X}))$ with $*$ denoting pointwise multiplication and $\bm{X}\oplus\bm{B} = \bm{X}^{\frac{1}{2}}\bm{B}\bm{X}^{\frac{1}{2}}$. Matrices $\bm{A}_{l}, \bm{B}_{l} \in \SPD$ are part of the network parameters $\bm{\theta_{\text{T, $\SPD$}}}$.

\textbf{Gyrospace hyperplane-based \ac{fc} \ac{spd} Layers (\ac{gyrospd}).} Inspired by their equivalent in hyperbolic spaces~\citep{shimizu2021hnn++}, another version of a gyrocalculus-based \ac{fc} layer for SPD networks was proposed by~\citet{nguyen2024mmn++}. Here, the classic combination of matrix product with a weight and addition of a bias is treated as hyperplane equation. Therefore, these fully-connected layers compute 
the distance to a hyperplane that is located in the origin of the manifold from a specific point $\bm{V}$, that is obtained from the layer input $\bm{X}^{(l-1)}\in \SPDn$ via the gyrocalculus formulation of the hyperplane equation
\begin{equation}
\label{eq:fclayer_hyperplane_equation}
    \bm{V}(\bm{X}^{(l-1)}) = \langle \ominus \bm{B}_{l} \oplus \bm{X}^{(l-1)}, \bm{A}_{l}\rangle,
\end{equation}
that is parametrized by the network parameters $\bm{A}_{l}, \bm{B}_{l} \in \SPDn$. The additive inverse $\ominus$ fulfills $\ominus \bm{X} = \bm{X}^{-1}$.
The final layer output is given by
\begin{equation}
    \label{eq:fclayer_output}
    \bm{X}^{(l)} = \text{exp}((\bm{X}^{(l-1)}) * \bm{I}_\text{v}),
\end{equation}
where $\bm{I}_{\text{v}(i,j)}=\frac{1}{\sqrt{2}}$ for $i\neq j$ and $\bm{I}_{\text{v}(i,j)}=1$ for $i=j$.

\textbf{ReEig Nonlinearities.} The ReEig layers were first introduced by \cite{huang2016riemannianspdnn}, and were extended to a more general context in \cite{Nguyen22gyromanifolds}. Even though $\SPD$ already classifies as nonlinear space, this type of layer adds the option of adding further nonlinearities to the network. The layer output is defined as  
\begin{equation}
    \label{eq:re_eig_nonlinearity}
    \bm{X}^{(l)} = \sigma_{\text{spd}}(\bm{X}^{(l-1)}) = \bm{U}\text{diag}(\text{max}(\epsilon \mathbf{I}, \bm{V}))\bm{U}^\trsp,
\end{equation}
with eigenvalue decomposition of $\bm{X}^{(l-1)} = \bm{U}\text{diag}(\bm{V})\bm{U}^\trsp$. The constant $\epsilon$ acts as rectification threshold to avoid smaller eigenvalues and hereby regularize the output.
Under the affine-invariant distance, this nonlinearity layer is analogous to a ReLu nonlinearity in the Euclidean network layers.

\section{Constrained AE: Implementation Details}
\label{appendix:constrained_ae}
The constrained AE is implemented following~\citep{Otto2023MOR}.
To guarantee the projection properties, the nonlinear activation functions employed in the encoder and decoder network $\sigma_{-}$ and $\sigma_{+}$ must satisfy $\sigma_{-} \circ \sigma_{+} = \text{id}$. To do so, they are defined by 
\begin{equation}
    \label{eq:AE_activation_functions}
    \sigma_{\pm}(x_i) = \frac{bx_i}{a}\mp \frac{\sqrt{2}}{a \sin(\alpha)} \pm \frac{1}{a}\sqrt{\left(\frac{2x_i}{\sin(\alpha)\cos(\alpha)}\mp\frac{\sqrt{2}}{\cos(\alpha)}\right) + 2a},
\end{equation}
with 
\begin{equation}
    \begin{cases}
        a &= \csc^2(\alpha) - \sec^2(\alpha),\\
        b &= \csc^2(\alpha) + \sec^2(\alpha).
    \end{cases}
\end{equation}
The activations then resemble smooth, rotation-symmetric versions of the common leaky ReLu activations. The parameter $0<\alpha<\frac{\pi}{4}$ sets the slope of the activation functions. Throughout our experiments, we set $\alpha = \frac{\pi}{8}$. 

For each layer $l$, the weight matrices are given by the pair $\left(\bm{\Phi}_l, \bm{\Psi}_l\right)\in \mathcal{B}_{n_l, n_{l-1}}$ (see App.~\ref{appendix:biorthogonal_manifold}). In the proposed \ac{rolnn}, we train these weights via Riemannain optimization on the biorthogonal manifold, analogous to the Riemannian optimization on the \ac{spd} manifold presented in Section~\ref{sec:model_training_riem_opt}. Notice that the optimization of the \ac{ae} parameters $\bm{\Xi}\!=\!\{\bm{\Phi}_{l}, \bm{\Psi}_{l}, \bm{b}_{l}\}_{l=1}^{L}$ takes place on a product of Euclidean and biorthogonal manifolds.

When learning dynamic latent space for Lagrangian \ac{mor}, the AE should learn the lifted embedding \eqref{eq:liftedembedding} and corresponding reduction maps. Therefore, the outputs of the encoder and decoder networks are differentiated with respect to their inputs. In our implementation, we take layerwise analytical derivatives and obtain the full differentials via the chain rule. A second derivative of the encoder and decoder networks enables training on acceleration data. Full-dimensional acceleration data is encoded via the reduction map $\rhoq$ as $\ddcheckbmq = d\rhoq\vert_{\bmq} \ddbmq + d^2\rhoq\vert_{(\bmq, \dbmq)} \dbmq$, and latent accelerations are reconstructed via the embedding $\varphiq$ as $\tilde{\ddbmq} = d\varphiq\vert_{\checkbmq} \ddcheckbmq + d^2\varphiq\vert_{(\checkbmq, \dcheckbmq)} \dcheckbmq$.
This requires the expensive computation of second-order derivatives, but is nevertheless part of our experiments.

Given a dataset of $T$ observations $\left\{\bmq_i, \dbmq_i, \ddbmq_i\right\}_{i=1}^{T}$, the AE is trained on a loss
\begin{equation}
    \label{eq:ae_loss_bio}
    \ell_{\text{cAE}} 
        = \frac{1}{N}\sum_{i=1}^N \|\tilde{\bmq}_i - \bmq_i \|^2 + \| \tilde{\dbmq}_i - \dbmq_i \|^2 + \|\tilde{\ddbmq}_i - \ddbmq_i\|^2,
\end{equation}
of position reconstructions $\tilde{\bmq}$, velocity reconstructions $\tilde{\dbmq}$, and acceleration reconstruction $\tilde{\ddbmq}$. If the velocity or acceleration encodings are not relevant, the respective terms can simply be removed.

We compare the performance of the constrained \ac{ae} trained via Riemannian optimization with the training scheme introduced in~\citep{Otto2023MOR}, where the biorthogonality of weights is achieved via the introduction of additional penalty losses. In this case, the overall loss only reaches a minimum value if the overparametrized weight matrices actually fulfill the biorthogonality conditions for each layer $l$.
The overparametrized constrained \ac{ae} loss is given by
\small
\begin{equation}
\label{eq:ae_loss_op_total}
    \ell_{\text{cAE, op}}  
        = \ell_{\text{cAE}} + \ell_{\text{cAE, reg}} \quad \text{with} \quad
        \ell_{\text{cAE, reg}} = \sum_{l=1}^{L} \|\bm{\Phi}^{(l)}_\text{e}\|_\text{F}^2+ \|\bm{\Phi}^{(l)}_\text{d}\|_\text{F}^2 + \|\bm{\Phi}^{(l)\intercal}_\text{e} \bm{\Phi}^{(l)}_\text{d} - \mathbf{I}\|_\text{F}^2 \|(\bm{\Phi}^{(l)\intercal}_\text{e} \bm{\Phi}^{(l)}_\text{d})^{-1}\|_\text{F}^2,
\end{equation} 
\normalsize
with the loss $\ell_{\text{cAE, reg}}$ regularizing the biorthogonality of the \ac{ae} weights. 
Note that this overparametrization does not guarantee the biorthogonality condition, in contrast to the Riemannian approach. In practice, the additional regularization loss $\ell_{\text{cAE, reg}}$ required an additional learning rate, resulting in more possible combinations of hyperparameters.

\section{Training on Acceleration}
\label{appendix:joint_training_acc}
\begin{figure}\centering\def\svgwidth{.8\linewidth}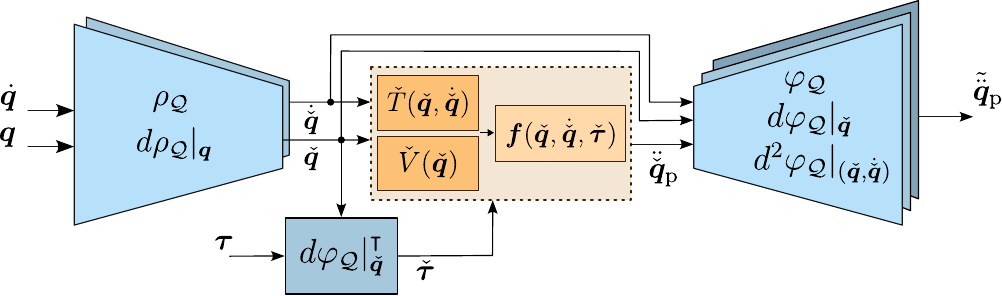
   \caption{Flowchart of the forward dynamics of the proposed reduced-order \ac{lnn} trained on acceleration via~\eqref{eq:lossROM_acc}. The reduction mappings and embeddings of the Lagrangian \ac{rom} are depicted in blue and parametrized via a constrained \ac{ae} with biorthogonal layers. The \ac{rom} dynamics are learned via a latent geometric \ac{lnn} depicted in orange. The mass-inertia matrix is parametrized via the \ac{spd} network described in Section~\ref{sec:SPD_LNNs}.}
   \vspace{-0.4cm}
   \label{fig:rom_drawing_acc}
\end{figure}

Fig.~\ref{fig:rom_drawing_acc} illustrates the complete forward dynamics of the proposed \ac{rolnn} trained on acceleration by minimizing the loss~\eqref{eq:lossROM_acc}.

%% file: figures/ROM_network2.pdf_tex
%% Creator: Inkscape 1.3.2 (091e20e, 2023-11-25), www.inkscape.org
%% Accompanies image file 'ROM_network2.pdf' (pdf, eps, ps)
\begingroup%
  \makeatletter%
  \providecommand\color[2][]{%
    \errmessage{(Inkscape) Color is used for the text in Inkscape, but the package 'color.sty' is not loaded}%
    \renewcommand\color[2][]{}%
  }%
  \providecommand\transparent[1]{%
    \errmessage{(Inkscape) Transparency is used (non-zero) for the text in Inkscape, but the package 'transparent.sty' is not loaded}%
    \renewcommand\transparent[1]{}%
  }%
  \providecommand\rotatebox[2]{#2}%
  \newcommand*\fsize{\dimexpr\f@size pt\relax}%
  \newcommand*\lineheight[1]{\fontsize{\fsize}{#1\fsize}\selectfont}%
  \ifx\svgwidth\undefined%
    \setlength{\unitlength}{480.61513345bp}%
    \ifx\svgscale\undefined%
      \relax%
    \else%
      \setlength{\unitlength}{\unitlength * \real{\svgscale}}%
    \fi%
  \else%
    \setlength{\unitlength}{\svgwidth}%
  \fi%
  \global\let\svgwidth\undefined%
  \global\let\svgscale\undefined%
  \makeatother%
  \begin{picture}(1,0.29463176)%
    \lineheight{1}%
    \setlength\tabcolsep{0pt}%
    \put(0,0){\includegraphics[width=\unitlength,page=1]{ROM_network2.pdf}}%
  \end{picture}%
\endgroup%

%% file: texfiles/experimental_appendix.tex
\section{Additional Experimental Details}
\label{appendix:additional_experimental details}
This section presents additional details on the experimental setup of Section~\ref{sec:experiments}. Position and velocity predictions are obtained via Euler forward integration.

\subsection{$2$-\ac{dof} Pendulum of Section~\ref{subsec:DoublePendulum}.}
\label{app:spd_abl_details}

\subsubsection{Simulation and Data Generation}
\label{appendix:spd_abl:sim_and_datasets}
For this experiment, we use a $2$-\ac{dof} pendulum implemented in \textsc{Mujoco}~\citep{Todorov12:mujoco}. The pendulum is connected via hinge joints. Both links $i$ are modeled as cylinders of radius $r_i = \SI{0.025}{\m}$, length $l_i = \SI{0.4}{\m}$, and mass $m_i = \SI{0.1}{\kg}$. The simulated environment is entirely free of dissipative effects.
We generate two different datasets $\mathcal{D}_{\text{pend}} = \{\{\bmq_{n,k}, \dbmq_{n,k}, \ddbmq_{n,k}, \bmtau_{n,k}\}_{k=1}^{K}\}_{n=1}^{N}$ consisting of $N$ trajectories with $K$ samples per trajectory. The state evolution is simulated in \textsc{Mujoco} with an RK4-solver with timestep $\Delta t = \SI{e-3}{\s}$ over a time interval $\mathcal{I} = [0, T] \SI{}{\second}$.

\textbf{Dataset 1: Unactuated pendulum.} The initial configurations are randomly sampled from the interval $q_i(t=0) \in \left[0, 30\right]\SI{}{\degree}$ for each \ac{dof} $i=\{1,2\}$. The initial velocities are set to $\dbmq(t=0) = \mathbf{0}$. Each trajectory is recorded for $T=2\SI{}{\s}$. All models presented in Section~\ref{subsec:DoublePendulum} in the main text are trained on this dataset.

\textbf{Dataset 2: Sine-Tracking pendulum.} Each joint is controlled to follow a reference trajectory via an inverse dynamics torque control law. The reference trajectories are set to follow sinusoidal shapes as 
$q_{\text{ref},i} = A_i \sin(2\pi f_i t + \phi)$, with $\phi_i = \arcsin\left(\frac{q_{i,0}}{A_i}\right)$, $q_{i,0} \in \left[0, 30\right]\SI{}{\degree}$, $A_i \in \left[1,30\right]\SI{}{\degree}$, $f_i\in\left[\frac{1}{15}, 1\right]\SI{}{\per\second}$, and $i=\{1,2\}$. $\dbmq_{\text{ref},i}$ and $\ddbmq_{\text{ref},i}$ are obtained as first and second derivatives of $q_{\text{ref},i}$ with respect to time. Initial positions and velocity are set to match the reference trajectories as $q_i(t=0) = q_{\text{ref},i}(t=0)$ and $\dot{q}_i(t=0) = \dot{q}_{\text{ref},i}(t=0)$
Each trajectory is recorded for $T=\SI{3.5}{\s}$. Results of models trained on this datasets are presented in App.~\ref{app:spd_abl_details_exp}.

\subsubsection{Architectures \& Training}
\label{appendix:spd_abl:architecture}
\textbf{Geometric \ac{lnn} architecture.}
For all models, the potential energy network $V(\bmq; \bm{\theta}_{\text{V}})$ consists of $L_{\text{V}}=2$ hidden Euclidean SoftPlus layers of $64$ neurons.
The kinetic energy of all models is parametrized via the mass-inertia matrix of the system. 
We consider several architectures for the mass-inertia matrix network $\bmM(\bmq; \bm{\theta}_{\text{T}})$, which differ in their components $g_{\mathbb{R}}$, $g_{\text{Exp}}$, and $g_{\SPD}$ as follows:
\begin{itemize}
    \item $g_{\mathbb{R}}$: The Euclidean component consists of $L_{\text{T}, \mathbb{R}}$ Euclidean layers on the tangent space of the SPD manifold with SoftPlus activations and $64$ neurons per layer.
    \item $g_{\text{Exp}}$: The basepoint $\bm{P}\in\SPD$ is either set as the origin $\mathbf{I}$, or learned as a network parameter $\bm{P}_{\bm{\theta}} \in \SPD$.
    \item $g_{\SPD}$: The SPD component consists of $L_{\text{T}, \SPD}$ hidden \ac{spd} layers. We consider two different \ac{fc} \ac{spd} layers, namely \ac{gyroai}~\eqref{eq:gyronl_layer}, and \ac{gyrospd}~\eqref{eq:fclayer_hyperplane_equation}. Both \ac{fc} layers can be augmented with a ReEig nonlinearity~\eqref{eq:re_eig_nonlinearity}. The regularization constant $\epsilon$ of the nonlinearity is uniformly set as $\epsilon=1\times 10^{-4}$. Note that for the SPD layers, the common notion of number of neurons per layer or layer width is not directly applicable. As the layers act as a transformation $\SPD \to \SPD$ via network parameters $\bm{\theta_{\text{T},\SPD}}$, they can be interpreted as carrying a single transformation per layer, or as transformations of manifold dimension $\text{dim}(\SPD) = \frac{n(n+1)}{2}$.
\end{itemize}

\textbf{Baseline \ac{lnn} architectures.}
We compare the proposed geometric \ac{lnn} against DeLaN, which parametrizes the mass-inertia matrix via a Cholesky network. We consider two variants here: \emph{(1)} The kinetic and potential energy networks are independent as in~\citep{Lutter2023DeLaN}. Both networks consist of $2$ Euclidean SoftPlus layers of $64$ neurons. Hereby, the potential energy network is fully equivalent to the ones employed in the geometric \ac{lnn}. The kinetic energy network returns a Cholesky decomposition $\bm{L}(\bm{q})$ of the mass-inertia matrix $\bmM(\bmq)$; \emph{(2)} The kinetic and potential energy networks share parameters, i.e., $\bm{\theta}_{\text{T}} \cap \bm{\theta}_{\text{V}}$ as in~\citep{Lutter19:DeLan}. The MLP consists of $2$ hidden Euclidean SoftPlus layers of $64$ neurons, and separate output layers return the potential energy and the Cholesky decomposition.

\textbf{Training on Acceleration Data.}
Each training dataset $\Dtrain$ consist of randomly-sampled data from $30$ trajectories from one of the datasets presented in Section~\ref{appendix:spd_abl:sim_and_datasets}. The models are trained by minimizing the \ac{lnn} loss~\eqref{eq:LNN_MSE} for $3000$ epochs with architecture-specific learning rates. 
If not further specified, both $g_{\mathbb{R}}$ and $g_{\SPD}$ consist of each 2 layers.

\textbf{Training on multi-step Integration.}
For these experiments, we randomly sample recordings $i$ form $30$ trajectories, which then form the training dataset 
$\Dtrain = \left\{\left\{\bmq\right\}_{t_{0,i}}^{t_{0,i}+H\Delta t}, \left\{\dbmq\right\}_{t_{0,i}}^{t_{0,i}+H\Delta t}, \left\{\bm{\tau}\right\}_{t_{0,i}}^{t_{0,i}+H\Delta t}\right\}$. Note that this training dataset depends on the ODE solver's timestep $\Delta t$ and desired prediction horizon $H$. The models are trained by minimizing the loss
\begin{equation} 
    \label{eq:fom_lnn_ode_loss}
        \ell_{\text{LNN}} = \frac{1}{HN}\sum_{i=1}^N \sum_{j=1}^H \int_{t_{i}}^{t_{i} + j\Delta t}\|\dcheckbmq_{\text{p}, i}(t_{i,j}) - \dcheckbmq_i(t_{i,j}) \|^2 + \|\tilde{\dbmq}_{\text{p}, i}(t_{i,j}) - \dbmq_{i}(t_{i,j})\|^2 + \gamma\; \|\bm{\theta}\|_2^2
\end{equation}
for $1500$ epochs with architecture-specific learning rates. Notice that the loss~\eqref{eq:fom_lnn_ode_loss} is the equivalent of~\eqref{eq:lossROM_ode} for \ac{fom}s. In this category, the featured kinetic energy networks all consist of $2$ hidden layers, and the exponential map layer is implemented with respect to $\mathbf{I}$.  

\textbf{Long-term prediction Experiments.} In those experiments, we use both models trained on acceleration data and with multi-step integration. In both scenario, the training datasets contain $|\Dtrain| = 8000$ samples and are trained for $2000$ epochs.

\subsection{Coupled Pendulum ($16$ \acp{dof}) of Section~\ref{subsec:coupled_pend}}
\label{appendix:coupled:details}
\subsubsection{Simulation and Data Generation}
\label{appendix:coupled:sim_and_dataset}
In this experiment, we consider a $16$-\ac{dof} coupled pendulum. The pendulum is connected via hinge joints. All links $i$ are modeled as capsules of radius $r_i = \SI{0.05}{\m}$, cylinder length $l_i = \SI{0.5}{\m}$, and mass $m_i = \SI{1}{\kg}$. The simulated environment is entirely free of dissipative effects. We generate a dataset $\mathcal{D}_{\text{pend16}} = \{\{\bmq_{n,k}, \dbmq_{n,k}, \ddbmq_{n,k}, \bmtau_{n,k}\}_{k=1}^{K}\}_{n=1}^{N}$ consisting of $N$ trajectories with $K$ samples per trajectory. 
The first $4$-\ac{dof} are simulated in \textsc{Mujoco}~\citep{Todorov12:mujoco} with an RK4-solver with timestep $\Delta t = \SI{e-3}{\s}$ over a time interval $\mathcal{I} = [0, 3] \SI{}{\second}$, while the last $12$ \acp{dof} are constrained to nonlinear combinations of the first $4$, as described in Table~\ref{tab:coupled:pendulum_dof_generation}.
The initial configurations of the first $4$ \acp{dof} are randomly sampled from the interval $q_i(t=0) \in \left[0, 30\right]\SI{}{\degree}$. The initial velocities are $\dbmq(t=0) = \mathbf{0}$.

\begin{table}[tbp]
\centering
\caption{Nonlinear combinations for \acp{dof} 5 to 16 of the coupled pendulum.}
\label{tab:coupled:pendulum_dof_generation}

\begin{tabular}{c|c}
\ac{dof} & $f(q_1, q_2, q_3, q_4)$ \\
\midrule
$q_5$ & $q_3 - \cos(q_2)$\\
$q_6$ & $q_1 + 0.1\sin(q_2)$\\
$q_7$ & $q_4 \cos(q_2)$\\
$q_8$ & $q_1 + q_3^2$\\
$q_9$ & $1.5\sin(q_2)$\\
$q_{10}$ & $-q_4 q_0$\\
$q_{11}$ & $\sin(q_1)$\\
$q_{12}$ & $0.4q_3q_4$\\
$q_{13}$ & $-0.9q_1 - q_2 + q_3 - 2q_{4}^{2}$\\
$q_{14}$ & $-3 \sin(q_3)$\\
$q_{15}$ & $-2 q_3^2$\\
$q_{16}$ & $-0.9 q_1^2$
\end{tabular}
\end{table}

\subsubsection{Architectures \& Training}
\label{appendix:coupled:architecture}
Our \ac{rolnn} is composed of a constrained \ac{ae} and a latent geometric \ac{lnn}. We consider a latent space of dimension $d=n_0=4$ to match the known underlying dimension of the high-dimensional pendulum dynamics.
The constrained \ac{ae} is implemented with $4$ biorthogonal encoder and decoder layers $\rhoq^{(l)}: \mathbbr{n_l} \to \mathbbr{n_{l-1}}$ and $\varphiq^{(l)}: \mathbbr{n_{l-1}} \to \mathbbr{n_{l}}$ of sizes $n_l = \left[8, 16, 16, 16\right]$. For the latent geometric \ac{lnn}, both the potential and kinetic energy network consist of $2$ hidden Euclidean layers of $64$ neurons. The kinetic energy network employs an exponential map layer with basepoint $\expmapblank{\mathbf{I}}$. 

We consider two versions of the \ac{rolnn}. The first model is trained on acceleration data by minimizing loss \eqref{eq:lossROM_acc}. The second model is trained with multi-step integration by minimizing \eqref{eq:lossROM_ode}. We consider $H_{\text{train}}=8$ latent integration steps and an integration time constant of $\SI{e-3}{\s}$. The losses are optimized using Riemannian Adam. For the first model, we use a learning rate of $5\times 10^{-2}$ for the \ac{ae} parameters $\bm{\Xi}$, $1\times 10^{-5}$ for the \ac{lnn} parameters $\bm{\theta}$, and a regularization $\gamma = 1\times 10^{-6}$. For the second model, the learning rate for the parameters of the \ac{lnn} is $2\times 10^{-4}$ with $\gamma=2\times 10^{-5}$. Both models are trained until convergence, i.e., for $4000$ and $3000$ epochs, respectively. 

We compare our \ac{rolnn} with \emph{(1)} a full-order geometric \acp{lnn}, \emph{(2)} L-OpInf~\citep{sharma24}, and \emph{(3)} L-Opinf with a latent LNN, as described in~\citep{sharma24NN}. 
For the full-order geometric \acp{lnn}, we only report results of the best-performing \ac{fom} out of $10$ models. The kinetic energy network consists of  $L_{\text{T},\mathbb{R}}=2$ Euclidean hidden SoftPlus layers with $128$ neurons, an exponential map layer with basepoint $\bm{P}=\mathbf{I}$, and no \ac{spd} layers, i.e., $L_{\text{T},\SPD}=0$. The potential energy network is likewise composed of $2$ Euclidean SoftPlus layers of $128$ neurons.
The full-order \ac{lnn} was trained for $3000$ epochs. The learning rate for the \ac{lnn} parameters $\bm{\theta}$ is set to $1\times 10^{-5}$, and $\gamma=1\times 10^{-5}$. 

For L-OpInf~\citep{sharma24}, we consider a projection onto a $4$-dimensional linear subspace, as for our \ac{rolnn}. We implemented L-OpInf in Python and solved the optimization problem leading to the reduced equations of motion with CVXPY. As described in~\citep{sharma24}, we set the low-dimensional mass-inertia matrix as a constant identity matrix, i.e., $\check{\bm{M}}=\mathbf{I}_d$. In our experiments, we optimize for $15000$ samples of ground-truth position data. 

For L-OpInf with latent LNN~\citep{sharma24NN}, we considered the same projection to a $4$-dimensional linear subspace as for the L-OpInf baseline. For fair comparisons, we enhance the latent LNN compared to that used in~\citep{sharma24NN}. In our experiment, the latent dynamic parameters are learned by a geometric LNN trained via the multi-step integration loss~\eqref{eq:lossROM_ode}, for which we use the same hyperparameters as for the RO-LNN trained with the same loss. Note that, for both versions of L-OpInf, the model is fed with ground truth data every $H_{\text{test}} = 25$ steps to maintain consistency in comparison with the \ac{rolnn}. We used the same Euler forward integration scheme as with our other models.

When investigating joint training in Section~\ref{subsec:experiments_lagrangian_rom}, the previously-described \ac{rolnn} trained with multi-step integration is compared to \emph{(1)} a version trained via the overparametrization of biorthogonal weights,  and \emph{(2)} a model that was trained separately, by sequentially training the \ac{ae} and latent \ac{lnn}. The overparametrized model uses the overparametrized loss~\eqref{eq:ae_loss_op_total}, with $\ell_{\text{cAE, reg}}$ multplied by a weighting constant $1\times 10^{-5}$. 
For the sequentially-trained model, the first $3000$ epochs were trained using only the AE components $\ell_{\text{AE}}$ of the loss~\eqref{eq:lossROM_ode}. For the consecutive $3000$ epochs, only the \ac{lnn} components $\ell_{\text{LNN}}$ were active of~\eqref{eq:lossROM_ode}. All models were trained on $|\Dtrain| = 24000$ samples, which were randomly sampled from $20$ trajectories. The testing dataset consists of $10$ trajectories.

\subsection{Rope ($192$ \acp{dof}) of Section~\ref{subsec:rope}}
\label{appendix:rope}
\subsubsection{Simulation and Data Generation}
\label{appendix:rope:sim_and_datasets}
The rope is implemented in \textsc{Mujoco} via an elastic cable. Over a total length of $\SI{1}{\m}$, the rope consists of $i=64$ equally-spaced capsule-shaped masses of $m_i=\SI{0.1}{\kg}$ and $r_i=\SI{0.02}{\m}$. Twisting and bending stiffness are set to $\SI{5e5}{\pascal}$, and the damping value of the joints is set to $0.1$. As the masses are connected via ball joints, the rope can move in all directions of the workspace.
For training and testing, we generate the datasets $\mathcal{D}_{\text{rope}} = \{\{\bmq_{n,k}, \dbmq_{n,k}, \bmtau_{n,k}\}_{k=1}^{K}\}_{n=1}^{N}$ consisting of $N$ trajectories with $K$ samples per trajectory. The generalized coordinates of the system are chosen to be the Cartesian positions $q_i = \left[x_i, y_i, z_i\right]$ of each mass's center of mass in the world frame.
To generate trajectories, we consider a scenario where one end of the rope is fixed to the origin and the other end is controlled to move along a circular trajectory of radius $r$. This mimicks a handheld manipulation task.

For each reference trajectory, a radius $r_i$ is sampled randomly from $r_i \in \left[0.05, 0.4\right]\SI{}{\m}$. To achieve non-planar motions, the circles are tilted by a random angle $\theta \in \left[-90, 90\right]\SI{}{\degree}$. The end of the rope is set to track such a reference at an angular velocity $\SI{120}{\degree\per\second}$ via a PD control.
The state evolution is simulated with an RK4-solver with timestep $\Delta t = \SI{e-3}{\s}$ over a time interval $\mathcal{I} = [0, 3.8] \SI{}{\s}$.

\subsubsection{Architectures \& Training}
\label{appendix:rope:architectures}
For our experiments on the rope, we consider four \acp{rolnn} that only differ from one another in the dimensions of the latent space. Each network is composed of a constrained \ac{ae} and a latent geometric \ac{lnn}. 
The constrained \ac{ae} is implemented with $4$ biorthogonal encoder and decoder layers $\rhoq^{(l)}: \mathbbr{n_l} \to \mathbbr{n_{l-1}}$ and $\varphiq^{(l)}: \mathbbr{n_{l-1}} \to \mathbbr{n_{l}}$ of sizes $n_l = \left[64, 64, 128, 192\right]$. We consider latent spaces of dimension $d = \left\{4,6,10,14\right\}$. In all scenarios, the kinetic energy network of the latent geometric \ac{lnn} consists of $2$ hidden Euclidean layers of $64$ neurons with SoftPlus activations, and an exponential map layer $\expmapblank{\mathbf{I}}$. The potential energy network also consists of $2$ hidden Euclidean layers of $64$ neurons with SoftPlus activations.

All models are trained on $H_{\text{train}}=8$ integration steps by minimizing the loss \eqref{eq:lossROM_ode} with Riemannian Adam~\citep{becigneul2018riemannianoptimization}. All models are trained with a learning rate of $8\times 10^{-2}$ for the \ac{ae} parameters $\bm{\Xi}$, $1\times 10^{-5}$ for the \ac{lnn} parameters $\bm{\theta}$, with a regularization $\gamma = 1\times 10^{-6}$. All models are trained until convergence for $3000$ epochs on $24000$ samples of training data, sampled randomly from $20$ different trajectories. For the testing dataset, we record $10$ full trajectories.

\subsection{Cloth ($600$ \acp{dof}) of Section~\ref{subsec:cloth}}
\label{appendix:cloth}
\subsubsection{Simulation and Data Generation}
\label{appendix:cloth:sim_and_datasets}
The deformable thin cloth is modelled in \textsc{Mujoco} as a flexible composite object with $200$ masses of $m_i = \SI{0.1}{\kg}$ equally spaced over a width of $\SI{0.1}{\m}$ and length of $\SI{0.2}{\m}$. For this model, the generalized coordinates are given by the Cartesian positions $q_i = \left[x_i, y_i, z_i\right]$ of each masses' center of mass in the world frame.

For training and testing, we generate datasets $\mathcal{D}_{\text{cloth}} = \{\{\bmq_{n,k}, \dbmq_{n,k}, \bmtau_{n,k}\}_{k=1}^{K}\}_{n=1}^{N}$ consisting of $N = 20$ trajectories during training, and $N=10$ trajectories during testing. The state evolution is simulated with an RK4-solver with timestep $\Delta t = \SI{e-4}{\s}$ over a time interval $\mathcal{I} = [0, 0.25] \SI{}{\s}$, resulting in $K=2500$ samples.
The trajectories contain recordings of the cloth falling on a sphere from a height of $\SI{0.13}{\m}$ in the center above the origin of the sphere. To vary scenarios, the radius of the sphere is randomly-sampled from $r \in \left[0.02, 0.12\right]\SI{}{\m}$.

\subsubsection{Architectures \& Training}
\label{appendix:cloth:architectures}
The \ac{rolnn} is composed of a constrained AE with layer sizes $n_l = \left[32, 64, 128, 600\right]$ for the $l = 1,...,4$ encoder and decoder layers $\rhoq^{(l)}: \mathbbr{n_l} \to \mathbbr{n_{l-1}}$ and $\varphiq^{(l)}: \mathbbr{n_{l-1}} \to \mathbbr{n_{l}}$, and a latent space dimension $n_0=10$. The kinetic energy network of the latent geometric \ac{lnn} consists of $2$ hidden Euclidean layers of $64$ neurons with SoftPlus activations, and an exponential map layer $\expmapblank{\mathbf{I}}$. The potential energy network also consists of $2$ hidden Euclidean layers of $64$ neurons with SoftPlus activations.

The featured model is trained on $H_{\text{train}}=8$ integration steps via the loss \eqref{eq:lossROM_ode} optimized using Riemannian Adam with a learning rate of $5\times 10^{-2}$ for the \ac{ae} parameters $\bm{\Xi}$, and $2\times 10^{-4}$ for the \ac{lnn} parameters $\bm{\theta}$, and a regularization $\gamma = 2\times 10^{-5}$. THe model is trained for $3000$ epochs.

\section{Additional Experimental Results}
\label{app:additional_results}
\subsection{Learning Lagrangian Dynamics with Geometric \acp{lnn}}
\label{app:spd_abl_details_exp}
This section presents additional results on learning the dynamics of a $2$-\ac{dof} pendulum with the geometric \ac{lnn} presented in Section~\ref{sec:SPD_LNNs}. 

\subsubsection{Additional Results on the Experiment of Section~\ref{subsec:DoublePendulum}}
\label{appendix:spd_abl:unactuated}
\begin{figure}[tbp]
    \centering
        \resizebox{\textwidth}{!}{
		\begin{tabular}{lllll}
            \choleskyblueline~Cholesky & \choleskyorangeline~Cholesky $\bm{\theta}_\text{T} \cap \bm{\theta}_\text{V}$& \expgreenline~$\expmapblank{\bm{I}}$ & \exppinkline~$\expmapblank{\bm{P}_{\bm{\theta}}}$ & \spdblueline~$\expmapblank{\bm{P}_{\bm{\theta}}}$,~\ac{gyrospd}
           \\ \spdredline~$\expmapblank{\bm{P}_{\bm{\theta}}}$,~\ac{gyrospd}+ReEig & \spdgreenline~$\expmapblank{\bm{P}_{\bm{\theta}}}$,~\ac{gyroai}+ReEig & \odeyellowline~$\expmapblank{\mathbf{I}},~H_\text{train}=1$ & \odepurpleline~$\expmapblank{\mathbf{I}},~H_\text{train}=8$ & \blackline~ Ground truth\\
        \end{tabular}
}
    \adjustbox{trim=0cm 0cm 0cm 0cm}{\includesvg[width=.9\linewidth]{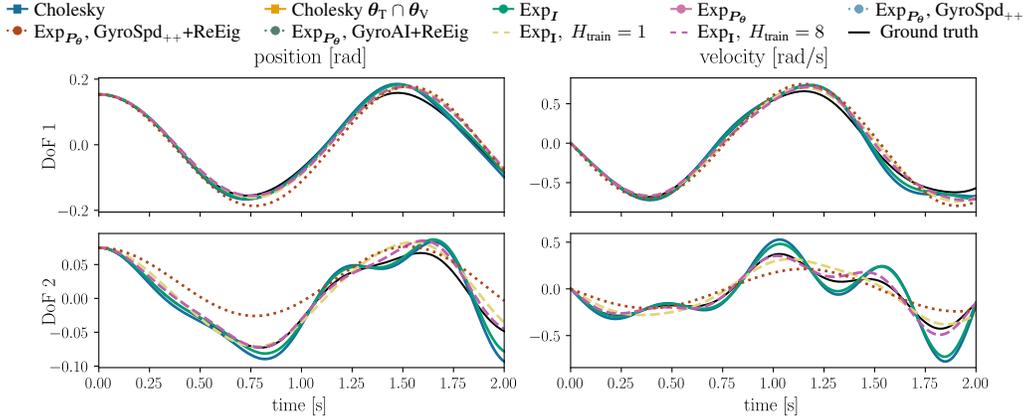}}
    \caption{Numerically-integrated position and velocity predictions and ground truth trajectory (\blackline) for \acp{lnn} trained on the unactuated $2$-\ac{dof} pendulum dataset. 
    }
    \label{fig:spd_abl:posvel_unact}
\end{figure}

\textbf{Trajectory prediction.} 
Fig.~\ref{fig:spd_abl:posvel_unact} complements Fig.~\ref{fig:spd_abl:test_error_over_set_size_unact}-\emph{middle, right}  by depicting the long-term position and velocity predictions of the different geometric \acp{lnn} considered in Section~\ref{subsec:DoublePendulum}. The velocity predictions follow the same trend as the positions: The geometric \acp{lnn} outperform the DeLaNs, the geometric \acp{lnn} trained via multi-step integration outperform those trained on acceleration data, and the geometric \acp{lnn} with \ac{spd} layers do not noticeably outperform those employing solely Euclidean and exponential map layers.

\begin{figure}[htb]
    \centering
        \resizebox{\textwidth}{!}{
		\begin{tabular}{lllll}
            \choleskyblueline~Cholesky & \choleskyorangeline~Cholesky $\bm{\theta}_\text{T} \cap \bm{\theta}_\text{V}$& \expgreenline~$\expmapblank{\bm{I}}$ & \exppinkline~$\expmapblank{\bm{P}_{\bm{\theta}}}$ & \spdblueline~$\expmapblank{\bm{P}_{\bm{\theta}}}$,~\ac{gyrospd}
           \\ \spdredline~$\expmapblank{\bm{P}_{\bm{\theta}}}$,~\ac{gyrospd}+ReEig & \spdgreenline~$\expmapblank{\bm{P}_{\bm{\theta}}}$,~\ac{gyroai}+ReEig & \odeyellowline~$\expmapblank{\mathbf{I}},~H_\text{train}=1$ & \odepurpleline~$\expmapblank{\mathbf{I}},~H_\text{train}=8$ & \blackline~Ground truth\\
        \end{tabular}
}
    \adjustbox{trim=0cm 0cm 0cm 0cm}{\includesvg[width=0.5\linewidth]{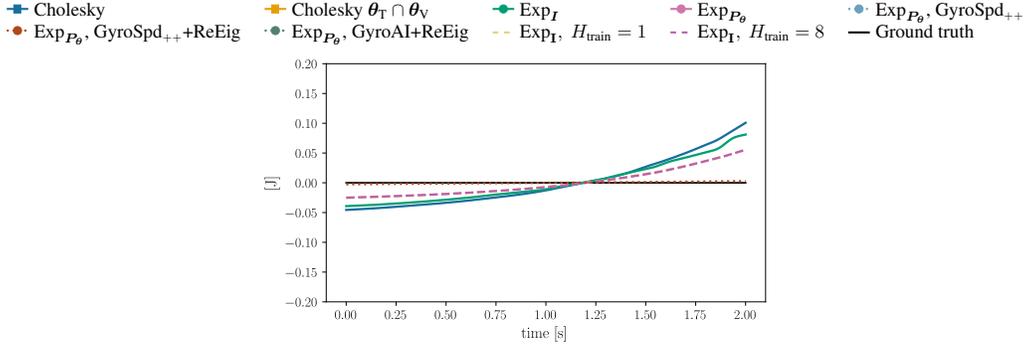}}
    \caption{Total energy $\check{\mathcal{E}}$ (up to a constant) of different \acp{lnn}.}
    \label{fig:spd_abl:energy_unact}
\end{figure}

\textbf{Energy conservation.}
Fig.~\ref{fig:spd_abl:energy_unact} displays the total energy 
of the different \acp{lnn} of Figs.~\ref{fig:spd_abl:test_error_over_set_size_unact}-\ref{fig:spd_abl:posvel_unact} along the predicted trajectory. As the pendulum is unactuated, the energy levels --- that can be learned up to a constant --- should be constant over the predicted trajectory. We observe that all \acp{lnn} approximately conserve the total predicted energy $\mathcal{E}$, with the DeLaNs showing the highest fluctuations. Note that, since the predicted trajectories are based on Euler forward integration, we do not expect the total energy of the system to be perfectly preserved.

\subsubsection{Sine-Tracking $2$-\ac{dof} Pendulum}
\label{appendix:spd_abl:sine_tracking}

We reproduce the experiment of Section~\ref{subsec:DoublePendulum} on a different dataset, namely on the sine-tracking $2$-\ac{dof} pendulum dataset introduced in App.~\ref{appendix:spd_abl:sim_and_datasets}.

\begin{figure}[t]
    \centering
    \resizebox{\textwidth}{!}{
		\begin{tabular}{lllll}
            \choleskyblueline~Cholesky & \choleskyorangeline~Cholesky $\bm{\theta}_\text{T} \cap \bm{\theta}_\text{V}$& \expgreenline~$\expmapblank{\bm{I}}$ & \exppinkline~$\expmapblank{\bm{P}_{\bm{\theta}}}$ & \spdblueline~$\expmapblank{\bm{P}_{\bm{\theta}}}$,~\ac{gyrospd}
           \\ \spdredline~$\expmapblank{\bm{P}_{\bm{\theta}}}$,~\ac{gyrospd}+ReEig & \spdgreenline~$\expmapblank{\bm{P}_{\bm{\theta}}}$,~\ac{gyroai}+ReEig & \odeyellowline~$\expmapblank{\mathbf{I}},~H_\text{train}=1$ & \odepurpleline~$\expmapblank{\mathbf{I}},~H_\text{train}=8$ &\\
        \end{tabular}
}

    \adjustbox{trim=0cm 0cm 0cm 0cm}{\includesvg[width=0.5\linewidth]{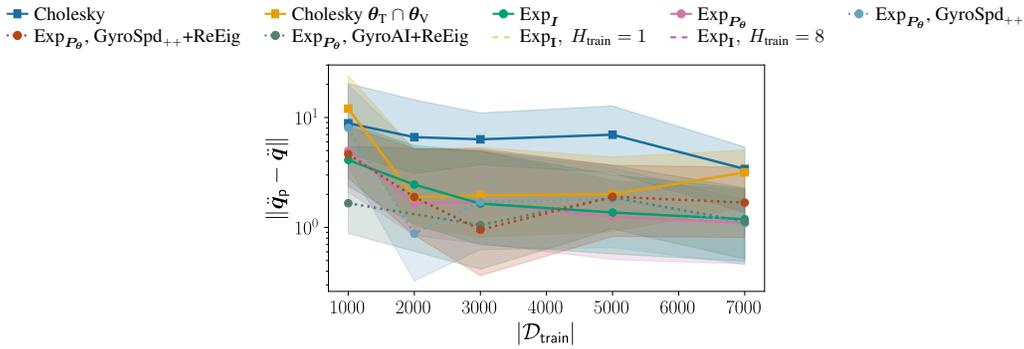}}
    \caption{Median acceleration prediction error for different \acp{lnn} and training set sizes $\Dtrain$ over $10$ testing trajectories. The models were trained on the sine-tracking $2$-\ac{dof} pendulum dataset. Shaded regions represent first and third quartiles. }
    \label{fig:spd_abl:test_error_over_set_size_sine}
\end{figure}

Fig.~\ref{fig:spd_abl:test_error_over_set_size_sine} depicts the acceleration prediction errors for selected architectures.
Similarly as for the unactuated dataset, the \acp{lnn} with Cholesky networks are outperformed by the geometric \acp{lnn}, and even more so in the low-data regime. In contrast to the unactuated pendulum dataset, the geometric \acp{lnn} featuring \ac{spd} layers result in slightly lower acceleration errors.

\begin{figure}[tbp]
    \centering
        \resizebox{\textwidth}{!}{
		\begin{tabular}{lllll}
            \choleskyblueline~Cholesky & \choleskyorangeline~Cholesky $\bm{\theta}_\text{T} \cap \bm{\theta}_\text{V}$& \expgreenline~$\expmapblank{\bm{I}}$ & \exppinkline~$\expmapblank{\bm{P}_{\bm{\theta}}}$ & \spdblueline~$\expmapblank{\bm{P}_{\bm{\theta}}}$,~\ac{gyrospd}
           \\ \spdredline~$\expmapblank{\bm{P}_{\bm{\theta}}}$,~\ac{gyrospd}+ReEig & \spdgreenline~$\expmapblank{\bm{P}_{\bm{\theta}}}$,~\ac{gyroai}+ReEig & \odeyellowline~$\expmapblank{\mathbf{I}},~H_\text{train}=1$ & \odepurpleline~$\expmapblank{\mathbf{I}},~H_\text{train}=8$ & \blackline~Ground truth\\
        \end{tabular}
}
    \adjustbox{trim=0cm 0cm 0cm 0cm}{\includesvg[width=.9\linewidth]{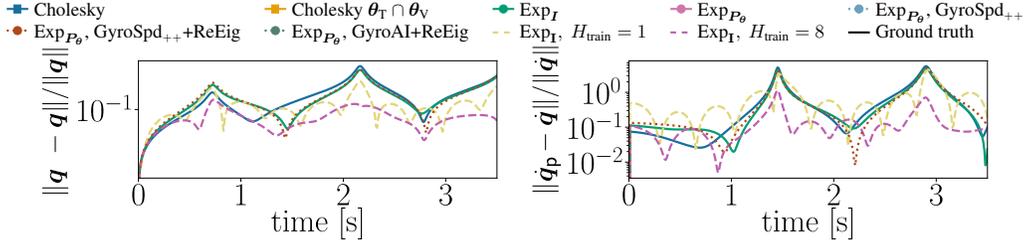}}
    \caption{Relative errors of numerically-integrated position and velocity predictions w.r.t. ground truth trajectory over prediction horizon $H_{\text{test}}$ for \acp{lnn} trained on the sine-tracking $2$-\ac{dof} pendulum dataset. 
    }
    \label{fig:spd_abl:posvel_sine}
\end{figure}

Fig.~\ref{fig:spd_abl:posvel_sine} depict long-term trajectory prediction obtained by integrating the state predictions given by the \ac{lnn}.
We observe that training the \acp{lnn} with multi-step integration is crucial to obtain accurate trajectory prediction for the sine-tracking dataset.
Generally, the results on this second dataset are aligned with those presented in Section~\ref{subsec:DoublePendulum} and validate the effectiveness of considering the geometry of the mass-inertia matrix and of training the models via multi-step integration when learning Lagrangian dynamics. 

\subsubsection{Influence of \ac{spd} layers}
\label{appendix:spd_abl:on_spd_man_nns}
Next, we evaluate the influence of \ac{spd} layers within the kinetic energy network $\bmM(\bmq; \bm{\theta}_{\text{T}})$ of the geometric \ac{lnn}. 
We consider the network architectures described in App.~\ref{appendix:spd_abl:architecture} and train the models with acceleration data. We vary the type of \ac{spd} layers, as well as the number of hidden layers $L_{\text{T},\mathbb{R}}$, $L_{\text{T},\SPD}$ of the Euclidean and \ac{spd} components of the kinetic energy network.
Each model is trained with two random seeds and tested over $10$ unseen testing trajectories from the corresponding dataset. We report the lower prediction error across the two trained models. 
The average acceleration prediction errors for the unactuated and sine-tracking datasets are given in Tables~\ref{tab:spd_abl:spd_layers_errors_unact} and~\ref{tab:spd_abl:spd_layers_errors_sine}, respectively.

Despite the increased performance of geometric \acp{lnn} over DeLaNs, the reported prediction errors do not allow us to observe any clear influence of the choice of \ac{spd} layers, exponential map basepoints, or Euclidean and \ac{spd} layers depths across training dataset sizes. In particular, increasing the depth of \ac{spd} layers does not seem to increase the expressivity of the overall \ac{spd} network. 

We hypothesize that the limited improvements obtained by adding the \ac{gyrospd}, \ac{gyroai}, and ReEig layers in our \ac{spd} network may be due to practical issues related to the training of these layers. Specifically, the abundance of matrix exponentials and logarithms within the \ac{spd} layers tends to cause numerical issues in the optimization procedure. These numerical instabilities also prevent the convergence of several models, as indicated by nan values in Tables~\ref{tab:spd_abl:spd_layers_errors_unact} and~\ref{tab:spd_abl:spd_layers_errors_sine}. Notice that this mostly occurs for the \ac{spd} networks containing ReEig layers.
Potential solutions include implementing analytic layer derivatives to replace unstable autodiff computations, or endowing the SPD manifold with a different Riemannian metric.

Due to the limited performance improvements and increased computational complexity (see App.~\ref{appendix:spd_abl:runtimes}) of the \ac{spd} layers), we consider \ac{spd} networks $\bmM(\bmq; \bm{\theta}_{\text{T}})= (g_{\text{Exp}}\circ g_{\mathbb{R}}) (\bmq)$ composed of Euclidean and exponential map components when learning reduced-order Lagrangian dynamics in the experiments of Section~\ref{subsec:experiments_lagrangian_rom}.

\begin{table}[tbp]
    \centering
    
    \caption{Mean and standard deviation of the acceleration prediction errors $\|\ddbmq_\text{p} - \ddbmq\|$ for geometric \acp{lnn} with different \ac{spd} network architectures trained on various training sizes on the unactuated $2$-\ac{dof} pendulum dataset.}
    \label{tab:spd_abl:spd_layers_errors_unact}
    \resizebox{\columnwidth}{!}{
    \begin{tabular}{c|c|c|c|c|c|c|c}
        \multicolumn{3}{c|}{} & \multicolumn{5}{c}{$|\Dtrain|$} \\
        \multicolumn{3}{c|}{} & $1000$ & $2000$ & $3000$ & $5000$ & $7000$ \\
        \hline
        \multirow{4}{*}{\ac{gyrospd}} & \multirow{2}{*}{$\text{Exp}_{\bm{I}}$} & $L_{\text{T},\mathbb{R}}=1, L_{\text{T},\SPD}=3$ & $5.95 \pm 5.69$ &  $5.22 \pm 3.78$ &  $1.83 \pm 2.61$ &  $1.86 \pm 2.98$ &  $\underline{0.92 \pm 1.31}$ \\
                                   &                                & $L_{\text{T},\mathbb{R}}=L_{\text{T},\SPD}=2$ & $15.7 \pm 19.5$ &  $5.15 \pm 3.66$ &  $4.43 \pm 2.96$ &  $2.71 \pm 3.22$ &  $1.52 \pm 1.69$ \\
        \cline{2-8}
                                   & \multirow{2}{*}{$\text{Exp}_{\bm{P}_{\bm{\theta}}}$} & $L_{\text{T},\mathbb{R}}=1, L_{\text{T},\SPD}=3$ & $\underline{1.78 \pm 2.30}$ &  $4.48 \pm 2.70$ &  $4.45 \pm 2.96$ &  $1.27 \pm 1.47$ &  $1.75 \pm 2.32$ \\
                                   &                                & $L_{\text{T},\mathbb{R}}=L_{\text{T},\SPD}=2$ & $4.37 \pm 2.65$ &  $5.34 \pm 3.72$ &  $2.08 \pm 3.02$ &  $\bm{0.89 \pm 1.11}$ &  $1.44 \pm 2.02$ \\
        \hline
        \multirow{4}{*}{\ac{gyrospd}+ReEig} & \multirow{2}{*}{$\text{Exp}_{\bm{I}}$} & $L_{\text{T},\mathbb{R}}=1, L_{\text{T},\SPD}=3$ & $4.66 \pm 2.99$ &  $1.75 \pm 2.33$ &  $\underline{1.77 \pm 2.40}$ &  $1.66 \pm 2.58$ &  $1.10 \pm 1.42$ \\
                                   &                                & $L_{\text{T},\mathbb{R}}=L_{\text{T},\SPD}=2$ & $4.66 \pm 2.81$ &  $1.72 \pm 2.19$ &  $1.80 \pm 2.52$ &  $1.78 \pm 2.21$ &  $\bm{0.84e \pm 1.25}$ \\
        \cline{2-8}
                                   & \multirow{2}{*}{$\text{Exp}_{\bm{P}_{\bm{\theta}}}$} & $L_{\text{T},\mathbb{R}}=1, L_{\text{T},\SPD}=3$ & $4.60 \pm 2.87$ & $4.58 \pm 2.93$ & $4.38 \pm 2.85$ & $1.43 \pm 1.89$ & $1.18 \pm 1.93$ \\
                                   &                                & $L_{\text{T},\mathbb{R}}=L_{\text{T},\SPD}=2$ & $1.99 \pm 2.23$ &  $1.78 \pm 2.38$ &  $1.79 \pm 2.52$ &  $\underline{1.20 \pm 1.77}$ &  $1.14 \pm 1.31$ \\
        \hline
        \multirow{4}{*}{\ac{gyroai}+ReEig} & \multirow{2}{*}{$\text{Exp}_{\bm{I}}$} & $L_{\text{T},\mathbb{R}}=1, L_{\text{T},\SPD}=3$ & $2.31 \pm 2.33$ &  $7.28 \pm 8.43$ &  $\bm{1.52 \pm 2.08}$ &  $1.32 \pm 1.91$ &  $1.13 \pm 1.48$ \\
                                   &                                & $L_{\text{T},\mathbb{R}}=L_{\text{T},\SPD}=2$ & $\bm{1.77 \pm 2.34}$ & nan & $1.99 \pm 2.70$ & nan & $1.30 \pm 2.17$ \\
        \cline{2-8}
                                   & \multirow{2}{*}{$\text{Exp}_{\bm{P}_{\bm{\theta}}}$} & $L_{\text{T},\mathbb{R}}=1, L_{\text{T},\SPD}=3$ & $2.76 \pm 2.49$ &  $\underline{1.71 \pm 2.29}$ &  $21.2 \pm 19.5$ &  $1.26 \pm 1.66$ &  $1.07 \pm 1.48$ \\
                                   &                                & $L_{\text{T},\mathbb{R}}=L_{\text{T},\SPD}=2$ & nan & $\bm{1.57 \pm 2.01}$ & nan & $1.82 \pm 2.48$ &  $4.22 \pm 2.60$ \\
    \end{tabular}
    }
\end{table}

\begin{table}[tbp]
    \centering    
    \caption{Mean and standard deviation of the acceleration prediction errors $\|\ddbmq_\text{p} - \ddbmq\|$ for different geometric \acp{lnn} with different \ac{spd} network architectures trained on various training sizes on the sine-tracking $2$-\ac{dof} pendulum dataset.}
    \label{tab:spd_abl:spd_layers_errors_sine}
    \resizebox{\columnwidth}{!}{
    \begin{tabular}{c|c|c|c|c|c|c|c}
        \multicolumn{3}{c|}{} & \multicolumn{5}{c}{$|\Dtrain|$} \\
        \multicolumn{3}{c|}{} & 1000 & 2000 & 3000 & 5000 & 7000 \\
        \hline
        \multirow{4}{*}{\ac{gyrospd}} & \multirow{2}{*}{$\text{Exp}_{\bm{I}}$} & $L_{\text{T},\mathbb{R}}=1, L_{\text{T},\SPD}=3$ & $\bm{3.73 \pm 4.46}$ &  $4.22 \pm 4.42$ &  $3.95 \pm 4.43$ &  $3.37 \pm 3.54$ &  $2.05 \pm 2.41$ \\
                                   &                                & $L_{\text{T},\mathbb{R}}=L_{\text{T},\SPD}=2$ & $10.9 \pm 8.62$ &  $3.95 \pm 4.50$ &  $3.89 \pm 4.49$ &  $2.80 \pm 2.66$ &  $2.18 \pm 2.29$ \\
        \cline{2-8}
                                   & \multirow{2}{*}{$\text{Exp}_{\bm{P}_{\bm{\theta}}}$} & $L_{\text{T},\mathbb{R}}=1, L_{\text{T},\SPD}=3$ & $12.0 \pm 13.6$ &  $\underline{3.58 \pm 4.58}$ &  $3.70 \pm 4.59$ &  $3.73 \pm 3.95$ & nan \\
                                   &                                & $L_{\text{T},\mathbb{R}}=L_{\text{T},\SPD}=2$ & $17.0 \pm 21.7$ &  $\bm{3.44 \pm 4.84}$ &  $3.66 \pm 4.40$ &  $2.75 \pm 2.80$ &  $\underline{1.49 \pm 1.47}$ \\
        \hline
        \multirow{4}{*}{\ac{gyrospd}+ReEig} & \multirow{2}{*}{$\text{Exp}_{\bm{I}}$} & $L_{\text{T},\mathbb{R}}=1, L_{\text{T},\SPD}=3$ & $4.59 \pm 4.58$ &  $4.17 \pm 4.43$ &  $4.15 \pm 4.48$ &  $71.0 \pm 79.6$ &  $5.80 \pm 5.23$ \\
                                   &                                & $L_{\text{T},\mathbb{R}}=L_{\text{T},\SPD}=2$ & $13.5 \pm 15.3$ &  $4.05 \pm 4.26$ &  $3.95 \pm 4.38$ &  $5.78 \pm 4.91$ &  $\bm{1.27 \pm 1.25}$ \\
        \cline{2-8}
                                   & \multirow{2}{*}{$\text{Exp}_{\bm{P}_{\bm{\theta}}}$} & $L_{\text{T},\mathbb{R}}=1, L_{\text{T},\SPD}=3$ & $4.29 \pm 4.53$ &  $4.55 \pm 4.47$ &  $4.13 \pm 4.43$ &  $\bm{1.88 \pm 1.74}$ &  $3.34 \pm 3.65$ \\
                                   &                                & $L_{\text{T},\mathbb{R}}=L_{\text{T},\SPD}=2$ & $6.13 \pm 4.61$ &  $3.90 \pm 4.42$ &  $3.45 \pm 4.81$ &  $2.95 \pm 3.17$ &  $2.64 \pm 2.82$ \\
        \hline
        \multirow{4}{*}{\ac{gyroai}+ReEig} & \multirow{2}{*}{$\text{Exp}_{\bm{I}}$} & $L_{\text{T},\mathbb{R}}=1, L_{\text{T},\SPD}=3$ & $6.39 \pm 5.73$ &  $4.00 \pm 4.45$ &  $\bm{3.33 \pm 4.30}$ & nan & $6.74 \pm 5.59$ \\
                                   &                                & $L_{\text{T},\mathbb{R}}=L_{\text{T},\SPD}=2$ & $9.14 \pm 9.73$ &  $3.79 \pm 4.41$ &  $3.67 \pm 4.60$ &  $2.36 \pm 2.58$ &  $49.1 \pm 58.2$ \\
        \cline{2-8}
                                   & \multirow{2}{*}{$\text{Exp}_{\bm{P}_{\bm{\theta}}}$} & $L_{\text{T},\mathbb{R}}=1, L_{\text{T},\SPD}=3$ & $4.02 \pm 4.66$ & nan & $4.11 \pm 4.34$ &  $\underline{2.35 \pm 2.08}$ &  $1.86 \pm 2.13$ \\
                                   &                                & $L_{\text{T},\mathbb{R}}=L_{\text{T},\SPD}=2$ & $\underline{3.99 \pm 4.66}$ & nan & $\underline{3.40 \pm 4.55}$ &  $2.78 \pm 2.68$ &  $1.68 \pm 1.75$ \\
    \end{tabular}
    }
\end{table}

\subsubsection{Influence of Multi-step Integration during Training}
\label{appendix:spd_abl:learn_from_horizon}
We assess the influence of the number $H_{\text{train}}$ of steps considered when training our geometric \acp{lnn}.
The featured networks are trained as described in~\ref{appendix:spd_abl:architecture}. For this experiment, the architectures are as described for multi-step integration in section~\ref{appendix:spd_abl:architecture}.
Table~\ref{tab:spd_abl:pred_horizon_influence} reports the velocity prediction errors averaged over $10$ testing trajectories for models trained on different prediction horizons $H_{\text{train}}$ for each of the $2$-\ac{dof} pendulum datasets.
We generally notice a significant decrease in the average error for geometric \acp{lnn} trained on longer horizons $H_{\text{train}}$. Ultimately, we select $H_{\text{train}}$ by trading off the increase in performance and the training time. The later is investigated next.

\begin{table}[tbp]
\centering
\caption{Mean and standard deviation of the $H_{\text{test}}=25$-steps velocity prediction errors for a geometric \ac{lnn} trained via multi-step integration with different horizons $H_{\text{train}}$ and datasets.}
\label{tab:spd_abl:pred_horizon_influence}
\renewcommand{\arraystretch}{1.0}
\resizebox{\textwidth}{!}{
    \begin{tabular}{c|cccc}
 Dataset&$H_{\text{train}}=1$ & $H_{\text{train}}=4$ & $H_{\text{train}}=8$ & $H_{\text{train}}=12$ \\
\midrule
Unactuated & $4.61\times 10^{-1} \pm 6.83\times 10^{-1}$ & $6.64\times 10^{-2} \pm 1.73\times 10^{-1}$ & $4.80\times 10^{-2} \pm 1.17\times 10^{-1}$ & $4.80\times 10^{-2} \pm 1.13\times 10^{-1}$\\
Sine-tracking &$2.32\times 10^{-2} \pm 4.07\times 10^{-2}$ & $8.87\times 10^{-1} \pm 2.05$ & $1.25\times 10^{-2} \pm 1.89\times 10^{-2}$ & $1.36\times 10^{-2} \pm 2.22\times 10^{-2}$\\
\end{tabular}}
\end{table}

\subsubsection{Runtimes}
\label{appendix:spd_abl:runtimes}
Table~\ref{table:spd_abl:times} shows the average training times of several \acp{lnn} trained on datasets of $|\Dtrain|=1000$ datapoints for $1000$ epochs. The potential energy network and Euclidean part of the kinetic energy network of all models consist of $2$ hidden Euclidean layers of $64$ neurons. For SPD layers, we only consider \ac{gyrospd}+ReEig layers as they are the computationally most demanding SPD layer combination. The featured model consists of $L_{\text{T}, \SPD} = 2$ of these layers.

Importantly, the well-performing geometric \ac{lnn} with identity basepoint and $L_{\text{T},\SPD}=0$ trains almost as fast as DeLaN with its Cholesky layers. Instead, integrating \ac{gyrospd}+ReEig layers within the \ac{spd} network doubles the training time. Note that similar increases were observed for other combinations of \ac{spd} layers.
As expected, training the \acp{lnn} with multi-step integration increases the training time. 
However, such training leads to increased data-efficiency and improved long-term predictions.
 
During evaluation, running the well-performing \ac{lnn} with $2$ Euclidean layers and the $\text{Exp}_{\mathbf{I}}$ layer in combination with numerical Euler integration over 25 steps, takes $0.01191$ secs, which could enable certain online applications.

\begin{table}[tbp]
		\centering
  \caption{Training times of various \acp{lnn} in seconds.}
  \label{table:spd_abl:times}
  \resizebox{.7\columnwidth}{!}{
		\begin{tabular}{c|c|c|c|c}
  \multicolumn{1}{c|}{DeLaN} & \multicolumn{4}{c}{Geometric \ac{lnn}} \\
  \midrule
Cholesky & $\expmapblank{\bm{I}}$ & $\expmapblank{\bm{P}_{\bm{\theta}}}$ & $\expmapblank{\bm{P}_{\bm{\theta}}}$, \ac{gyrospd}+ReEig  & $\expmapblank{\bm{I}}$, $H_{\text{train}}=8$ \\
\bottomrule
$88$&$92$&$121$&$189$&$105$\\
        \end{tabular}
        }
	\end{table}

\subsection{Additional Results on the $16$-\ac{dof} Coupled Pendulum Experiment of Section~\ref{subsec:coupled_pend}}
This section presents additional results on learning the high-dimensional dynamics of a $16$-\ac{dof} pendulum with geometric \ac{rolnn}.

\subsubsection{Acceleration vs Multi-step Training}
\vspace{-0.1cm}
\label{appendix:coupled:latent_pred_horizon}
Here, we further evaluate the impact of the losses used to train geometric \acp{rolnn}. Table~\ref{tab:pend:kstep_pred_architectures} shows the average acceleration errors of trajectory predictions of \acp{rolnn} trained on accelerations or via multi-step integration with horizon $H_{\text{train}} = 8$.
We observe that the \ac{rolnn} trained via multi-step integration consistently outperforms the \ac{rolnn} trained on accelerations for all prediction horizons $H_{\text{test}}$. This validates the effectiveness of considering successive states during training to learn the dynamics of physical systems. 
Notice that training on acceleration data requires computing the second derivatives in the \ac{ae} layers, resulting in large-scale matrix multiplications, which not only leads to longer training times, but depending on dimensionality of the system, requires vast amounts of memory. In this sense, training via multi-step integration also avoids the computation of these Hessian matrices.

\begin{table}[tbp]
    \centering
    \caption{Mean and standard deviations of the reconstructed and latent prediction errors for different prediction horizons with \acp{rolnn} trained on acceleration and via multi-step integration. }
    \label{tab:pend:kstep_pred_architectures}
    \resizebox{.98\textwidth}{!}{
    \begin{tabular}{c|c|c|c|c}
        \multicolumn{2}{c|}{} & $\sum_{j=1}^{H}\|\tilde{\bmq}_\text{p}(t_{j}) - \bmq(t_{j}) \|^2$ & $\ell_{\text{LNN},n}$ & $\sum \ell_{\text{LNN},d}$  \\
        \hline
        \multirow{2}{*}{$H_{\text{test}}=8$} & acc & $2.65\times 10^{-1} \pm 3.13\times 10^{-1}$& $1.95\times 10^{0} \pm 2.14\times 10^{0}$ &  $1.70\times 10^{-3} \pm 4.94\times 10^{-2}$ \\
        
                                   & ODE &  $\bm{1.86\times 10^{-4} \pm 2.57\times 10^{-4}}$ & $\bm{4.06\times 10^{-2} \pm 8.70\times 10^{-2}}$ &  $\bm{9.22^{-4} \pm 3.76\times 10^{-3}}$  \\
        \hline
        \multirow{2}{*}{$H_{\text{test}}=25$} & acc & $2.64\times 10^{-1} \pm 3.12\times 10^{-1}$ & $2.25\times 10^{0} \pm 2.50\times 10^{0}$ & $1.45\times 10^{-2} \pm 4.28\times 10^{-2}$  \\
                                   & ODE & $\bm{2.05\times 10^{-4} \pm 2.74\times 10^{-4}}$& $\bm{1.66\times 10^{-1} \pm 4.85\times 10^{-1}}$ & $\bm{7.78\times 10^{-3} \pm 3.25\times 10^{-2}}$  \\
        \hline
        \multirow{2}{*}{$H_{\text{test}}=50$} & acc & $2.64\times 10^{-1} \pm 3.10\times 10^{-1}$& $3.26\times 10^{0} \pm 4.62\times 10^{0}$ & $5.31\times 10^{-2} \pm 1.753\times 10^{-1}$ \\
                                   & ODE & $\bm{4.03\times 10^{-4} \pm 8.62\times 10^{-4}}$& $\bm{5.41\times 10^{-1} \pm 1.70\times 10^{0}}$ & $\bm{2.76\times 10^{-2} \pm 1.16\times 10^{-1}}$  \\
    \end{tabular}
    }
    \vspace{-0.2cm}
\end{table}

\subsubsection{Constrained vs Standard Autoencoders}
\label{appendix:coupled:ae_pos_rec_comparison}
\vspace{-0.1cm}
We further investigate the effectiveness of our geometric constrained \ac{ae} whose parameters are obtained via Riemannian optimization compared to two baselines, namely \emph{(1)} constrained \ac{ae} optimized via the overparametrization of~\citep{Otto2023MOR}, and \emph{(2)} a regular (unconstrained) \ac{ae}.
All \acp{ae} are trained to reduce a set of position data $\Dtrain = \left\{\bmq_i\right\}_{i=1}^{N}$ from $15$ trajectories of the $16$-\ac{dof} pendulum to a $d=4$-dimensional latent space.
The constrained \acp{ae} are implemented as in Section~\ref{subsec:coupled_pend} and described in App.~\ref{appendix:coupled:architecture}. 
The regular \ac{ae} consists of $4$ Euclidean layers with ReLu activations of sizes $n_l = \left[64,64,64,16\right]$ for $l = 1,...,4$ encoder and decoder layers $\rhoq^{(l)}: \mathbbr{n_l} \to \mathbbr{n_{l-1}}$ and $\varphiq^{(l)}: \mathbbr{n_{l-1}} \to \mathbbr{n_{l}}$.
Our constrained \ac{ae} is trained using Riemannian Adam with learning rate $5\times 10^{-2}$. The overparametrized and regular \acp{ae} are trained using Adam with learning rate $5\times 10^{-2}$ for the overparametrized, and $1e{-2}$ for the regular \ac{ae}. All models are trained for $2500$ epochs.

\begin{figure}[tbp]
    \centering
    \adjustbox{trim=0cm 0cm 0cm 0cm}{\includesvg[width=\linewidth]{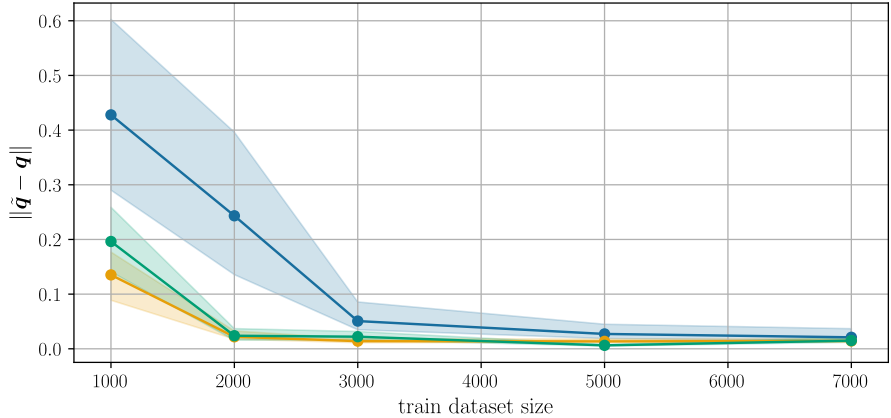}}
    \caption{Median and quartiles of position reconstruction error for the regular AE (\blueoneline), the biorthogonal constrained AE (\orangeline), and the overparametrized constrained AE (\greenline) for \ac{rolnn} trained on different sizes of training sets.}
    \label{fig:coupled_pend:ae_pos}
\end{figure}

Fig.~\ref{fig:coupled_pend:ae_pos} shows the reconstruction errors of the \acp{ae} for different sizes. We observe that the constrained \acp{ae} consistently outperform the regular one. Moreover, the Riemannian optimization leads to lower reconstruction errors than the overparametrization, especially in low-data regimes.

\subsubsection{Computational Speedup of ROM w.r.t. FOM}
\label{appendix:coupled:computational_comparison_rom_fom}
\vspace{-0.1cm}
Despite that we consider scenarios where the FOM is unknown and that trainings of a full-order LNN were unsuccessful (see Section \ref{subsec:coupled_pend}), we aim at providing an idea of the computational effort of our model compared to the evaluation of the FOMs. To do so, we symbolically derive the Lagrangian equations of motion with known physical quantities of the considered 16-DoF pendulum from Section \ref{appendix:coupled:sim_and_dataset} to obtain a FOM. We compare the wall-clock time of evaluation of this FOM against the ODE-trained RO-LNN. For both models, we consider a roll-out of $3000$ timesteps from the same initial conditions $\bmq(t=\SI{0}{\s})$ and $\dbmq(t=\SI{0}{\s})$ via Euler forward integration with a timestep $\Delta t = \SI{e-3}{\s}$. Averaged over 10 runs on the same local CPU, we achieve evaluation times of $\SI{113.59}{\s}$ for the FOM, and $\SI{1.57}{\s}$ for the ROM. We conclude that the ROMs obtained with our proposed method not only enable structure-preserving learning of high-dimensional system dynamics, but also significantly improve computational efficiency compared to evaluating FOMs.

\subsection{Additional Results on the $192$-\ac{dof} Rope Experiment of Section~\ref{subsec:rope}}
\label{appendix:rope:results}

\begin{figure}[tbp]
    \centering
    \includesvg[width=\linewidth]{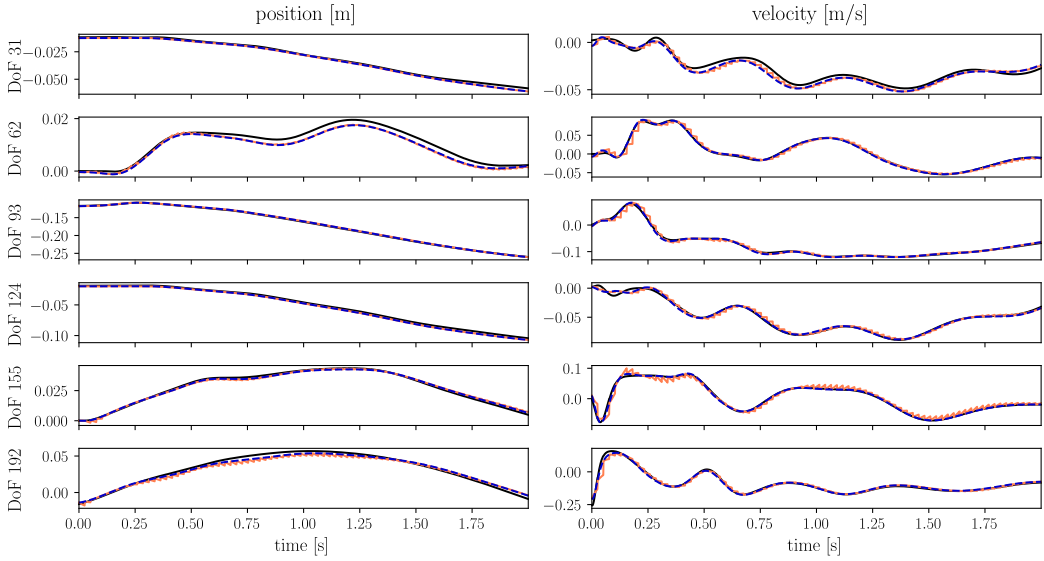}
    \caption{Rope position and velocity predictions (\orangeline) from the \acp{rolnn} trained via multi-step integration and ground truth (\blackline). The corresponding \ac{ae} reconstructions (\darkblueline) are depicted for completeness. The model is updated with a new initial condition $(\bmq_0, \dbmq_0)$ every $\SI{0.025}{s}$ ($H_{\text{test}}=25$).}
    \label{fig:rope:k_step_traj}
\end{figure}

Fig.~\ref{fig:rope:k_step_traj} shows the \ac{ae} reconstruction and the \ac{rolnn} predictions ($H_{\text{test}}=25$) for selected \acp{dof} of a test trajectory. We observe that the the \ac{rolnn} accurately models the high-dimensional dynamics of the rope.

\subsection{Additional Results on the $600$-\ac{dof} Cloth Experiment of Section~\ref{subsec:cloth}}
\label{appendix:cloth:results}

\begin{table}[t]
\centering
\caption{Cloth reconstruction and $H_{\text{test}}=25$-steps prediction errors over $10$ testing trajectories for the \ac{rolnn} with $d=10$ trained via multi-step integration with the loss~\eqref{eq:lossROM_ode}.  
}
\label{tab:cloth:loss_components}
\resizebox{.8\textwidth}{!}{
\begin{tabular}{cccc}
$\|\tilde{\bmq} - \bmq \|^2$ & $\|\tilde{\dbmq} - \dbmq \|^2$ & $\ell_{\text{LNN}, n}$ & $\ell_{\text{LNN}, d}$\\[0.2pt]
\midrule
$2.47\times 10^{-3} \pm 1.15\times 10^{-3}$ & $1.50\times 10^{0} \pm 1.81\times 10^{0}$&
 $1.51\times 10^{0} \pm 1.81\times 10^{0}$&$5.07\times 10^{-4} \pm 1.09\times 10^{-3}$\\
\end{tabular}
}
\end{table}

\begin{figure}[t]
    \centering
    \adjustbox{trim=0cm 0cm 0cm 0cm}{\includesvg[width=\linewidth]{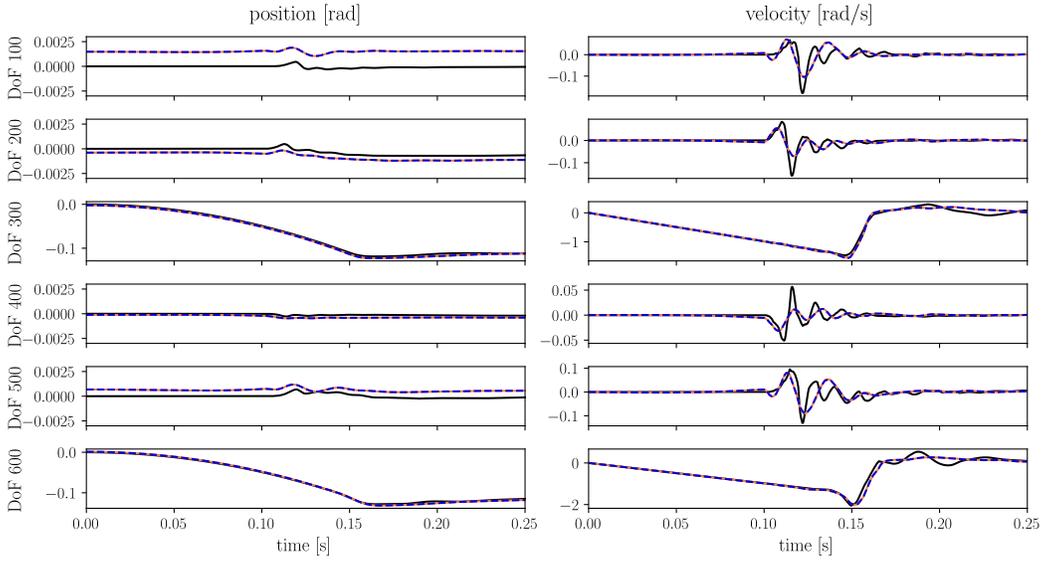}}
    \caption{Cloth position and velocity predictions (\orangeline) from the \acp{rolnn} trained via multi-step integration and ground truth (\blackline). The corresponding \ac{ae} reconstructions (\darkblueline) are depicted for completeness. The model is updated with a new initial condition $(\bmq_0, \dbmq_0)$ every $\SI{0.025}{s}$ ($H_{\text{test}}=25$). }
    \label{fig:cloth:k_step_pred}
\end{figure}

\begin{figure}[t]
    \centering
    \adjustbox{trim=0cm 0cm 0cm 0cm}{\includesvg[width=\linewidth]{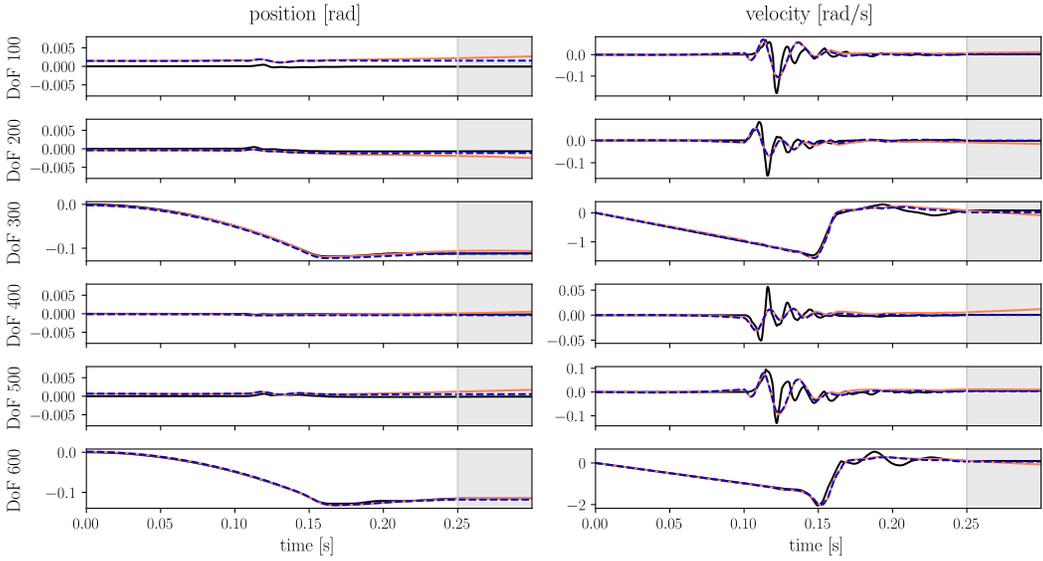}}
    \caption{Cloth position and velocity full-horizon predictions (\orangeline) from the \acp{rolnn} trained via multi-step integration and ground truth (\blackline). The corresponding \ac{ae} reconstructions (\darkblueline) are depicted for completeness. The dynamics are predicted from a given initial condition $(\bmq_0, \dbmq_0)$ for $\SI{0.3}{s}$ ($H_{\text{test}}=3000$). The grey-shaded areas indicate the interval during which ground-truth data are extrapolated from the last observation.}
    \label{fig:cloth:longterm_pred}
\end{figure}

\begin{figure}[t]
    \centering
    \includesvg[width=\linewidth]{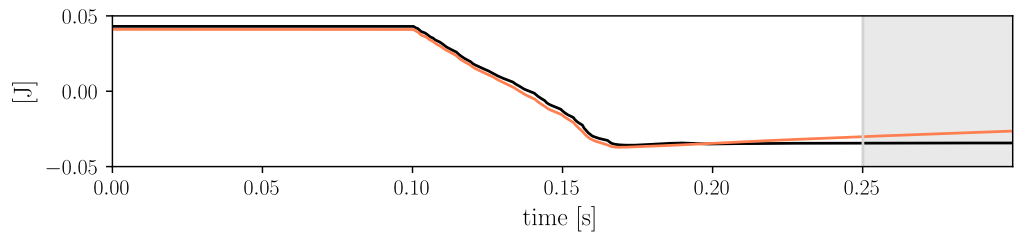}
    \caption{Predicted (\orangeline) and ground truth (\blackline) latent energy $\check{\mathcal{E}}$ (up to a constant) over time. The grey-shaded areas indicate the interval during which ground-truth data are extrapolated from the last observation.}
    \label{fig:cloth:energy_pred}
\end{figure}

Table~\ref{tab:cloth:loss_components} shows the reconstruction and prediction errors of our model averaged over $10$ testing trajectories.
Moreover, Figs.~\ref{fig:cloth:k_step_pred} and~\ref{fig:cloth:longterm_pred} show the \ac{ae} reconstruction and the predictions of the \ac{rolnn} of selected \acp{dof} of a test trajectory for horizons $H_{\text{test}}=25$ and $H_{\text{test}}=3000$, respectively.
Notice that the ground truth only contains actuation data for the first $\SI{0.25}{s}$ (i.e., $2500$ timesteps). Therefore, we reasonably assume that the cloth remains still for the additional time horizon $t_\text{aug} = \left[0.25, 0.30\right]\SI{}{\s}$ and set the ground truth states and torques as equal to their last recorded values.
Fig.~\ref{fig:cloth:energy_pred} displays the predicted latent energy to be compared with the groud-truth energy projected in the latent space. Overall, our results demonstrate the ability of our \ac{rolnn} to infer long-term predictions of complex high-dimensional deformable systems.

